\pdfoutput=1
\documentclass[11pt]{article}
\usepackage[letterpaper,margin=1in]{geometry}
\usepackage[T1]{fontenc}
\usepackage{amsmath}
\usepackage{newtxtext}
\usepackage{newtxmath}
\usepackage{graphicx}
\usepackage{booktabs}
\usepackage{xcolor}
\usepackage{fvextra}
\usepackage[numbers,sort&compress]{natbib}
\usepackage[hidelinks]{hyperref}
\usepackage[font=small,labelfont=bf,labelsep=period]{caption}
\input{glyphtounicode}
\hypersetup{pdftitle={Unmentioned Checklist Findings Change How Reinforcement Learning Appears to Improve Chest Radiograph Report Checking},pdfauthor={Ali Vosoughi, Akhil Kasturi, Chenliang Xu, Axel Wismueller}}
\renewcommand{\figurename}{Fig.}

\begin{document}
\begin{center}
{\LARGE\bfseries Unmentioned Checklist Findings Change How Reinforcement Learning Appears to Improve Chest Radiograph Report Checking\par}
\vspace{1.2em}
{\large Ali Vosoughi\textsuperscript{1,2,$\ast$}, Akhil Kasturi\textsuperscript{2}, Chenliang Xu\textsuperscript{1}, Axel Wismueller\textsuperscript{3,4,2}\par}
\vspace{0.8em}
{\small
\textsuperscript{1}Department of Computer Science, University of Rochester, Rochester, NY 14627, USA.
\textsuperscript{2}Department of Electrical and Computer Engineering, University of Rochester, Rochester, NY 14627, USA.
\textsuperscript{3}Department of Imaging Sciences, University of Rochester Medical Center, Rochester, NY 14642, USA.
\textsuperscript{4}Department of Biomedical Engineering, University of Rochester, Rochester, NY 14627, USA.\\[0.3em]
\textsuperscript{$\ast$}Corresponding author: Ali Vosoughi (ali.vosoughi@rochester.edu).\\[0.3em]
ORCiD: A.V. 0000-0003-1014-2937; A.K. 0009-0002-1079-2606; C.X. 0000-0002-2183-822X; A.W. 0009-0005-2048-1555.\par}
\end{center}

\begin{abstract}
\noindent Automated checks of radiology reports may rely on AI-generated checklists that leave findings unmentioned. We used reinforcement learning to train a vision-language model to fill in a 12-finding checklist from a chest radiograph without seeing the sentence under test; a separate checking model judged the sentence from the checklist. On held-out patients, a rule-based check and an independent medical checker, neither used in training, measured discrimination gains (Youden index) of 12.6\% and 11.8\%; only the rule-based check met the prespecified false-alarm criterion. Switching to the training format, which fixes finding order and enters unmentioned findings as absent, raised the training checker's measured gain and lowered the independent checker's, a prespecified comparison that yielded 6.2\% (95\% interval 2.0\% to 10.5\%) and, post hoc on held-out patients, 7.7\%. Across 8 checking models, acceptance of a label-consistent negative statement about an unmentioned finding ranged from 1.0\% to 97.0\%. Labels were report-derived, not radiologist-adjudicated.
\end{abstract}

\section*{Introduction}

\begin{figure}[t]
  \centering
  \includegraphics[width=0.66\linewidth]{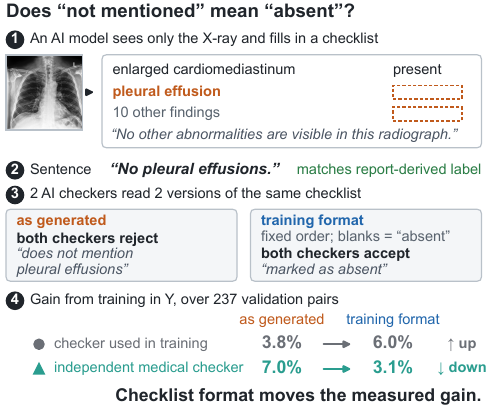}
  \caption{\textbf{Does ``not mentioned'' mean ``absent''?} Without seeing the report sentence, an AI model fills in a checklist of 12 possible findings from 1 radiograph, shown here in short with its free-text note for 1 illustrative CheXpert Plus study. The checklist has no entry for pleural effusion, although its general note says ``No other abnormalities are visible in this radiograph.'' The sentence ``No pleural effusions.'' agrees with the study's report-derived label. Both checking models reject the sentence on the checklist as generated and accept it in the training format, which puts the findings in a fixed order and enters unmentioned findings as absent. Because the general note can itself be read as excluding effusion, the example shows a missing finding-specific entry rather than complete silence. Across 237 matched validation pairs, switching to the training format moves the gain from training in pair-level discrimination ($Y$) from 3.8\% to 6.0\% for the checker used in training and from 7.0\% to 3.1\% for an independent medical checker.}
  \label{fig:teaser}
\end{figure}

Artificial intelligence is increasingly used to check statements in radiology reports against the images they describe \cite{mahmood2023factcheck,rextrust2024}. One design separates describing the image, often finding by finding, from deciding whether a report statement is supported by that description, within 1 model or across 2 models \cite{mj1_2026}. Such descriptions, like structured reports, rarely address every finding. For a radiologist, an explicit negative statement such as ``no pleural effusion'' documents an absence, whereas silence about effusion establishes neither presence nor absence, and structured templates prompt the reader to document relevant negatives \cite{kahn2009reporting,esr2018structured,schwartz2011structured}. A checking model has no such convention unless it is given one, so it must decide for itself whether a finding the description does not mention counts as absent.

The decision matters most when the image-reading model is trained. Vision-language models are now improved with reinforcement learning, in which a reward computed by another model scores their output \cite{lambert2024tulu,deepseekmath2024,guo2025deepseek,liu2025visualrft,huang2025visionr1,shen2025vlmr1,pan2025medvlmr1}, and image descriptions have been trained in this way for readers that never see the image \cite{fisch2020capwap,caprl2025,caprlpp2026,tang2026cccaption}. If the checking model that computes the training reward and the checking model that later evaluates the result treat unmentioned findings differently, the same description earns different credit from each, and part of a measured improvement reflects the checker rather than the image reading. Neither checker can be assumed to be neutral: vision-language models answer from language priors rather than the image \cite{agrawal2018dont,goyal2017making,tong2024eyes,chen2024rightway} and describe objects that are not there \cite{rohrbach2018object,li2023evaluating}, and model checkers are swayed by the wording and presentation of what they read \cite{mmjudgebias2026,modalens2026,verificationmirage2026,zheng2023judging,chen2024mllmjudge,sclar2024quantifying,imgtextinvariance2026,judgeconstruct2026,judgeharness2026}. In radiology, generated reports are usually scored with label extractors, entity graphs, and learned metrics \cite{smit2020chexbert,jain2021radgraph,yu2023evaluating,green2024}, and rewards and verifiers have mostly targeted the report writer \cite{mahmood2023factcheck,radprm2025,rextrust2024}, whereas here we train the description a checker relies on.

We studied this question in a blinded 2-step design for frontal chest radiographs (Figs.~\ref{fig:teaser} and~\ref{fig:method}). A checklist model sees the radiograph, but never the report sentence under test, and fills in a checklist of 12 findings; a checking model then sees the checklist and the sentence, but never the image, and decides whether the sentence is supported. Keeping the sentence out of the image-reading step, as in a blinded read, prevents the sentence from shaping the description, an idea shared with blind re-captioning \cite{blindrecap2026} and with checkers that commit to an answer before reading the claim \cite{commitfirst2026}. A concurrent study found that no checklist it prompted made the verdict depend on the image more than reading the sentence first \cite{anonconcurrent2027}, so we trained the checklist model instead. Training used matched pairs of studies from CheXpert Plus whose report-derived labels agree on every finding except the one a report sentence describes, so that the sentence is consistent with the label of one study and inconsistent with the label of the other, and each checklist was rewarded when the checking model's verdict agreed with the label. Unlike question-conditioned methods \cite{visionsr1_2025,vepa2026,moca2026}, the model being trained never sees the sentence, but a model-computed reward can still be exploited instead of improving at the task \cite{amodei2016concrete,skalse2022defining,gao2023scaling}, so every saved checklist was re-evaluated by 3 checkers: the checking model used in training, an independent medical checking model, and a rule-based check that uses no language model.

In a replication on 3 fresh training runs, whose evaluation protocol was recorded after the training jobs were launched but before any of their checklists was read, the rule-based check and the independent checker measured gains in discrimination, as a pair-level Youden index, of 12.6\% and 11.8\% on 273 held-out pairs from unseen patients, although the prespecified bound on false alarms was not established for the independent checker. During training the checking model read every checklist with its unmentioned findings entered as absent, whereas the trained checklist models list 7.6, 10.7, and 11.9 of the 12 findings on average. Reading the same saved checklists in the training format, with the findings in a fixed order and those absences entered, moved the training checker's estimate of the gain from 3.8\% to 6.0\% and the independent checker's from 7.0\% to 3.1\%; the prespecified comparison of the 2 checkers yielded a difference of 6.2\% $[2.0\%, 10.5\%]$, which a post hoc re-evaluation of held-out patients repeated at 7.7\% $[2.8\%, 12.8\%]$ (Fig.~\ref{fig:teaser}). Post hoc rereads with 6 further checking models showed that checkers differ widely in whether they accept a negative statement that agrees with the report-derived label about a finding the checklist does not mention, while a prespecified photograph study with single-object statements detected no comparable difference. Reading the sentence first with chain-of-thought reasoning discriminated better in point estimate than the trained checklists, so we treat the checklist as evidence whose measured value depends on the checker, not as a better verdict.

\section*{Results}

\subsection*{A blinded 2-step design trained on matched pairs}

\begin{figure*}[t]
  \centering
  \includegraphics[width=\linewidth]{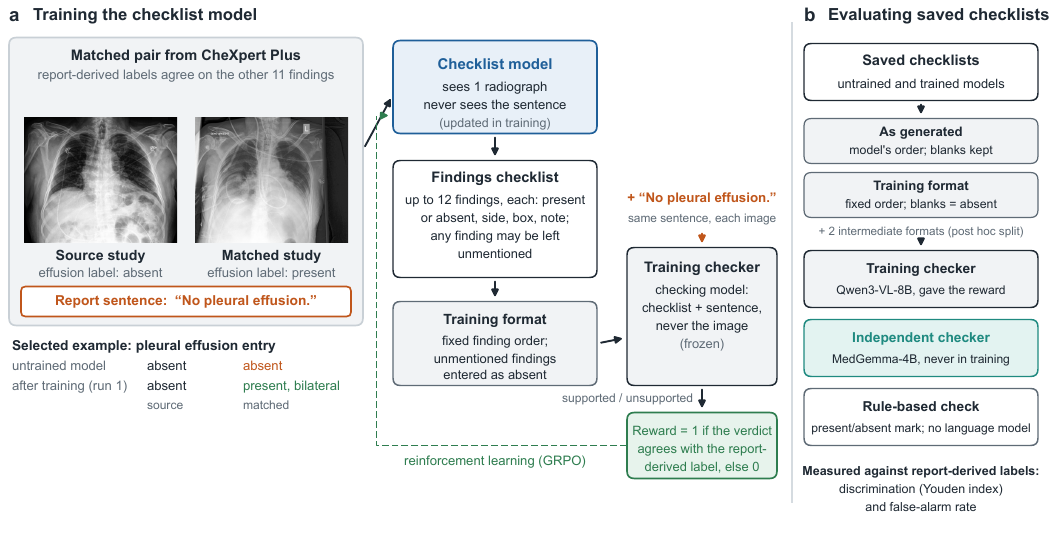}
  \caption{\textbf{Study design.} (a) Training. A matched pair is 2 frontal studies whose report-derived labels agree on 11 findings and differ on the finding a report sentence describes; here the sentence ``No pleural effusion.'' is consistent with the label of the source study (effusion absent) and inconsistent with the label of the matched study (effusion present). Each image is processed separately. The checklist model sees only the image and fills in a 12-finding checklist, which is rewritten in the training format, a fixed finding order with every unmentioned finding entered as absent. A frozen checking model reads that checklist and the sentence, never the image, and the checklist model is rewarded when the verdict agrees with the report-derived label (reinforcement learning by group relative policy optimization, GRPO). Pairs are matched on report-derived labels, not on radiographic appearance. Burned-in technologist initials on the 2 radiographs were masked; side and position markers were kept. The rows below the images show the effusion entry of this pair's checklists before training and after training in run 1; the pair was chosen by a rule fixed in advance, the first effusion pair whose trained checklists agree with the report-derived labels on both images under the rule-based check and the independent checker. A selected example illustrates the mechanism and does not show typical improvement. (b) Evaluation. The saved checklists of the untrained and trained models are read as generated and in the training format, with 2 intermediate formats that separate the fixed order from the entered absences, by the checking model used in training, by an independent medical checking model never used in training, and by a rule-based check of the present or absent mark that uses no language model.}
  \label{fig:method}
\end{figure*}

\begin{table}[t]
  \centering\footnotesize
  \renewcommand{\arraystretch}{1.15}
  \begin{tabular}{p{0.24\linewidth}p{0.70\linewidth}}
    \toprule
    Term & Meaning in this paper \\
    \midrule
    Checklist model & The vision-language model that looks at 1 frontal radiograph and fills in the checklist. It never sees the report sentence under test and is the only model that is trained. \\
    Checklist & Up to 12 listed findings, each entry giving present or absent, side, an optional box, and a short note. Findings may be left unmentioned. \\
    Checker & Any of the 3 methods that give a verdict on a sentence from a checklist: the training checker, the independent checker, and the rule-based check. \\
    Checking model & A frozen language model that reads only the checklist and 1 report sentence and answers supported or unsupported. The training checker supplied the reward; the independent checker, MedGemma, never did. \\
    Rule-based check & Supports the sentence when the checklist's present or absent mark for the sentence's finding agrees with the sentence. It reads the repaired checklist, counting an unmentioned finding as absent, and uses no language model. \\
    Report-derived label & Present or absent for each finding, extracted automatically from the report text by the CheXpert labeler, not a radiologist's reading of the image. \\
    Matched pair & 2 frontal studies whose labels agree on 11 findings and differ on the finding a report sentence describes, so the sentence agrees with the label of the source study and contradicts the label of the matched study. \\
    As generated; training format & The checklist as the model wrote it, in its own order with blanks kept; or rewritten with the findings in a fixed order and every unmentioned finding entered as absent, the format read in training. Entering an absence adds a statement the model did not make. \\
    Discrimination, $Y$ & Pair-level Youden index: sensitivity plus specificity minus 1, where sensitivity is acceptance of the sentence on source images and specificity is its rejection on matched images. A checker that ignores the image scores 0. \\
    False alarm, FA & Rejection of a sentence on the source image, whose label it agrees with; the false-alarm rate equals 1 minus sensitivity. \\
    Prespecified; post hoc & Written with its decision rule to a dated project record before the data it governs were read; or declared after related results had been read. \\
    \bottomrule
  \end{tabular}
  \caption{Terms used in this paper. Every measure is computed against report-derived labels, not radiologist reads of the images.}
  \label{tab:terms}
\end{table}

The design has 2 steps (Fig.~\ref{fig:method}, Table~\ref{tab:terms}). The checklist model receives a frontal chest radiograph and a fixed instruction, never the report sentence, and fills in a checklist of 12 findings, each entry stating whether the finding is present, its side, an optional box, and a short note; it may leave findings unmentioned. The checking model receives the checklist and a report sentence, never the image, and answers supported or unsupported, so any influence of the image on the verdict has to pass through the checklist. During training the checking model stays frozen. Each checklist earns a reward of 1 when the checker's verdict on its image's report sentence agrees with the report-derived label and 0 otherwise, and group relative policy optimization uses these rewards to update the checklist model, which never sees the sentence. The 2 studies of a pair match on their other report-derived labels and differ on the label the sentence addresses, so a checklist that ignores the image cannot tell them apart and scores 0 on the pair-level Youden index $Y$. We also report the false-alarm rate FA, because rejecting more sentences can raise $Y$. Every saved checklist is re-read by the checker used in training (the training checker), by MedGemma, a medically adapted model of another family that was never used in training (the independent checker), and by the rule-based check (Methods).

\subsection*{Training improved discrimination on unseen patients, but the false-alarm criterion was not met}
\label{sec:res_replication}

\begin{table}[t]
  \centering\small
  \setlength{\tabcolsep}{4pt}
  \begin{tabular}{lcccccc}
    \toprule
    Checker & Untrained $Y$ & Untrained FA & $\Delta Y$ {\scriptsize[95\%]} & $\Delta$FA {\scriptsize(upper 95\%)} & Gain & False alarms \\
    \midrule
    \multicolumn{7}{l}{\emph{Validation pairs, 237}} \\
    Rule-based check & 3.0 & 45.1 & $+$8.3 {\scriptsize[$+$2.0, $+$14.1]} & $+$1.7 {\scriptsize($+$6.6)} & met & not met \\
    Independent checker & 2.1 & 36.3 & $+$7.0 {\scriptsize[$+$0.4, $+$13.3]} & $+$4.4 {\scriptsize($+$9.6)} & met & not met \\
    Training checker & 4.6 & 46.0 & $+$3.8 {\scriptsize[$-$1.0, $+$8.2]} & $+$12.0 {\scriptsize($+$15.9)} & \multicolumn{2}{c}{not in criterion} \\
    \midrule
    \multicolumn{7}{l}{\emph{Held-out test pairs, 273}} \\
    Rule-based check & 3.3 & 40.7 & $+$12.6 {\scriptsize[$+$6.9, $+$18.2]} & $-$2.7 {\scriptsize($+$1.9)} & met & met \\
    Independent checker & 2.2 & 35.9 & $+$11.8 {\scriptsize[$+$5.2, $+$18.6]} & $+$3.4 {\scriptsize($+$8.8)} & met & not met \\
    Training checker & 4.0 & 49.1 & $+$6.6 {\scriptsize[$+$2.3, $+$11.0]} & $+$6.8 {\scriptsize($+$11.2)} & \multicolumn{2}{c}{not in criterion} \\
    \bottomrule
  \end{tabular}
  \caption{Training improved discrimination for both checkers not used in training, but the joint criterion on discrimination and false alarms failed in both populations. Replication on the validation pairs and confirmation on held-out test pairs from unseen patients, all entries in percent; checklists read as generated. The protocol was recorded after the 3 training jobs were launched and before any of their checklists was read; the runs differ only in their random seed. $\Delta Y$ is the run-averaged paired change in discrimination from the untrained model, with its two-sided 95\% interval over patient groups (per-run values in the Supplementary Information), and $\Delta$FA the averaged change in the false-alarm rate with its one-sided 95\% upper bound. The criterion asked each checker not used in training for $\Delta Y \geq 5\%$ with the lower bound above zero (gain) and an upper bound on $\Delta$FA of at most $5\%$ (false alarms). A change in a rate is the difference of the 2 percentages. Intervals are conditional on the 3 trained models.}
  \label{tab:main}
\end{table}

The training recipe was developed on the validation pairs, where the original criterion was never met (Supplementary Information), so the replication, whose protocol was recorded after its training jobs were launched but before any of their checklists was read, asked 1 question: do 3 fresh training runs of 40 steps produce checklists that the checkers not used in training credit, without exceeding a margin on false alarms? Each endpoint was evaluated once on the validation pairs and once on the held-out test pairs (Table~\ref{tab:main}).

Both checkers not used in training measured the gain the criterion asks for in both populations, and every run is positive under both (Table~\ref{tab:main}).

The false-alarm condition was not met, its one-sided upper bound staying within the $5\%$ margin only for the rule-based check on the held-out test pairs. The results therefore do not establish that the rise in false alarms stayed within the margin, nor that it exceeded it, and the protocol recorded before any replicated checklist was read classifies this outcome in both populations as a trade-off. On the held-out test pairs 1 run carries the independent checker's rise, $+9.9\%$ against $+1.8\%$ and $-1.5\%$, the run whose checklists leave the most findings unmentioned (next section).

The training checker, whose verdicts supplied the reward, reads the same checklists differently. Its false-alarm rate rises by $12.0\%$ and $6.8\%$ for gains of $3.8\%$ and $6.6\%$, and on the validation pairs its rise in false alarms exceeds the rule-based check's by $10.3\%$ $[+5.8\%, +14.8\%]$ and the independent checker's by $7.6\%$ $[+3.3\%, +11.9\%]$, in prespecified contrasts, while its smaller gain differs from theirs by unresolved amounts.

\subsection*{Switching to the training format moved the 2 checkers' measured gains in opposite directions}
\label{sec:res_interface}

\begin{table}[t]
  \centering\scriptsize
  \setlength{\tabcolsep}{2.5pt}
  \begin{tabular}{lcccc}
    \toprule
    & \multicolumn{2}{c}{Training checker} & \multicolumn{2}{c}{Independent checker} \\
    Format of untrained and trained checklists & $\Delta Y$ & $\Delta$FA & $\Delta Y$ & $\Delta$FA \\
    \midrule
    As generated (evaluation default) & $+$3.8 {\scriptsize[$-$1.0, $+$8.2]} & $+$12.0 {\scriptsize[$+$7.6, $+$16.5]} & $+$7.0 {\scriptsize[$+$0.4, $+$13.3]} & $+$4.4 {\scriptsize[$-$1.4, $+$10.5]} \\
    Fixed order, unmentioned left out & $+$3.7 {\scriptsize[$-$1.0, $+$8.2]} & $+$12.0 {\scriptsize[$+$7.4, $+$16.6]} & $+$6.2 {\scriptsize[$-$0.3, $+$12.8]} & $+$4.9 {\scriptsize[$-$0.9, $+$10.9]} \\
    Model's order, absences entered & $+$5.8 {\scriptsize[$+$0.8, $+$10.8]} & $+$6.4 {\scriptsize[$+$1.4, $+$11.5]} & $+$5.4 {\scriptsize[$-$1.0, $+$12.0]} & $+$0.8 {\scriptsize[$-$4.5, $+$6.5]} \\
    Training format (fixed order, absences entered) & $+$6.0 {\scriptsize[$+$1.0, $+$10.8]} & $+$6.9 {\scriptsize[$+$1.9, $+$11.9]} & $+$3.1 {\scriptsize[$-$3.7, $+$9.5]} & $+$1.8 {\scriptsize[$-$3.9, $+$7.9]} \\
    \midrule
    Checker difference, training format (prespecified) & \multicolumn{4}{c}{$+$6.2 {\scriptsize[$+$2.0, $+$10.5]} on 237 pairs} \\
    of which the entered absences (post hoc) & \multicolumn{4}{c}{$+$4.6 {\scriptsize[$+$0.7, $+$8.6]} on 236 pairs} \\
    \bottomrule
  \end{tabular}
  \caption{The 2 checkers respond in opposite directions to the training format. The same saved validation checklists of the 3 replication runs, read in 4 formats, all entries in percent. Each row applies 1 format to untrained and trained checklists alike, with run-averaged changes in discrimination ($\Delta Y$) and the false-alarm rate ($\Delta$FA) and 95\% intervals over patient groups, on 237 pairs for the as-generated and training-format rows and 236 for the 2 intermediate rows. The difference between checkers is the independent checker's change from the training format to the checklists as generated minus the training checker's; it was prespecified before any replicated checklist was read. Its split into the fixed order and the entered absences (the 2 middle rows and the last row) was declared after the first run's checklists had been read as generated and in the training format, before the 2 intermediate formats were read, and is post hoc. The rule-based check is unchanged by every row.}
  \label{tab:form}
\end{table}

The checklist models often leave findings unmentioned although prompted to cover all 12, listing 10.4 on average before training, with at least 1 finding unmentioned on 59\% of checklists, and 7.6, 10.7, and 11.9 after it (Supplementary Information). During training each checklist was read in the training format (Methods), whereas every checker in Table~\ref{tab:main} read it as generated, so before any replicated checklist was read we prespecified the difference between the 2 checkers in how the training format changes their measured gain. Entering an absence is not a neutral change of display, since it adds an explicit statement the checklist model did not make.

The change of format moves the 2 checkers' point estimates in opposite directions (Fig.~\ref{fig:evidence}a). In the training format the training checker's gain rises from $3.8\%$ to $6.0\%$ and the independent checker's falls from $7.0\%$ to $3.1\%$, and the prespecified comparison of the 2 checkers yielded a difference of $6.2\%$ $[+2.0\%, +10.5\%]$ (Table~\ref{tab:form}), although neither checker's own change is resolved, and it holds without the 7 pairs whose checklists the format also repairs (Supplementary Information). In a post hoc analysis of the 236 pairs read in all 4 formats, the entered absences account for $4.6\%$ $[+0.7\%, +8.6\%]$ of that difference, against an unresolved $1.6\%$ $[-0.4\%, +3.6\%]$ for the fixed order, and they reduce the rise in false alarms by $5.4\%$ $[+2.6\%, +8.0\%]$ for the training checker and $3.3\%$ $[+0.6\%, +6.2\%]$ for the independent checker. In the full training format the training checker's rise falls from $12.0\%$ to $6.9\%$, so unmentioned findings explain part of its false alarms, not all, and fewer false alarms need not mean a better checker, since an entered absence can also support a negative statement on the matched image that its label contradicts.

The effect repeats on unseen patients. Re-evaluating the saved held-out test checklists in the same formats, declared post hoc as those rereads were launched and before any was analysed, gives a difference between the checkers of $7.7\%$ $[+2.8\%, +12.8\%]$ on 270 held-out test pairs, with an absence component of $6.4\%$ $[+2.2\%, +10.6\%]$. There the training checker's gain rises from $6.4\%$ to $11.2\%$ $[+6.8\%, +16.1\%]$ and its rise in false alarms falls from $7.2\%$ to $2.0\%$, so the handling of unmentioned findings carries most of that rise, although the prespecified outcome, defined on the checklists as generated, stands.

Entering unmentioned findings as absent lowers the estimated rise in false alarms under every checking model and instruction we read, and telling the checkers that an unlisted finding is absent does not remove the difference, $7.4\%$ $[3.1\%, 12.1\%]$ on the validation pairs and $5.0\%$ $[-0.4\%, 10.7\%]$ on the held-out test pairs (Supplementary Information).

\subsection*{Checking models disagree on what an unmentioned finding means}
\label{sec:res_silence}

\begin{table}[t]
  \centering\scriptsize
  \setlength{\tabcolsep}{1.6pt}
  \begin{tabular}{lccc}
    \toprule
    & Negative accepted & \multicolumn{2}{c}{Difference from the training checker} \\
    Checker & (finding unmentioned) & Validation & Held-out \\
    \midrule
    Training checker & 64.0 / 53.5 & -- & -- \\
    MedGemma-4B & 79.0 / 74.0 & $+$6.2 {\scriptsize[$+$2.0, $+$10.5]} & $+$7.7 {\scriptsize[$+$2.8, $+$12.8]} \\
    Gemma-3-4B & 1.0 / 1.6 & $+$0.6 {\scriptsize[$-$4.0, $+$4.8]} & $+$4.9 {\scriptsize[$-$0.2, $+$10.2]} \\
    Gemma-3-12B & 91.0 / 91.3 & $+$1.5 {\scriptsize[$-$1.6, $+$4.7]} & $+$3.2 {\scriptsize[$+$0.1, $+$6.3]} \\
    Gemma-3-27B & 56.0 / 50.4 & $+$0.3 {\scriptsize[$-$2.4, $+$3.0]} & $+$1.9 {\scriptsize[$-$1.1, $+$5.0]} \\
    \midrule
    \multicolumn{2}{l}{Both, quote-the-entry instruction} & $+$2.8 {\scriptsize[$-$2.9, $+$8.5]} & $+$3.2 {\scriptsize[$-$2.2, $+$9.4]} \\
    \multicolumn{2}{l}{Both, reason-first instruction} & $+$1.8 {\scriptsize[$-$2.7, $+$6.3]} & $-$1.4 {\scriptsize[$-$6.9, $+$4.1]} \\
    \bottomrule
  \end{tabular}
  \caption{Checking models disagree on whether an unmentioned finding counts as absent. All entries in percent. Negative accepted: verdicts accepting a negative statement that agrees with its report-derived label, such as no pleural effusion, about a finding the checklist does not mention (100 validation / 127 held-out test verdicts). Difference: the checker's format effect minus the training checker's (Table~\ref{tab:form}), 95\% intervals over patient groups, on 221 to 274 pairs; in the last 2 rows both primary checkers read under an instruction to quote the checklist entry or to reason first. MedGemma's differences are those of Table~\ref{tab:form}; all other entries come from post hoc rereads of the same saved checklists. Phi-4-mini, OLMo-2-7B, and Qwen3-VL-30B-A3B, which complete the 8 checkers, are reported in the Supplementary Information.}
  \label{tab:judges}
\end{table}

Under the default instruction, checking models vary widely in accepting a negative statement that agrees with its report-derived label, such as ``no pleural effusion'', about a finding the checklist does not mention. In rereads of the same saved checklists by 6 further checking models and under 2 instructions, all declared post hoc before they were made, such a statement is accepted by $1.0\%$ of Gemma-3-4B's verdicts \cite{gemma3_2025}, $64.0\%$ of the training checker's, $79.0\%$ of MedGemma's, and $97.0\%$ of Phi-4-mini's on the validation checklists (Table~\ref{tab:judges}). On those checklists, entering unmentioned findings as absent lifts every checker but Gemma-3-4B to $97.0\%$ to $100\%$ acceptance of these label-consistent negative statements and to $89.9\%$ to $97.5\%$ of label-inconsistent ones, whose finding is labelled present but unmentioned, raising acceptance of both alike. MedGemma also accepts more positive statements that the filled checklist contradicts, label-inconsistent positive statements about an unmentioned finding rising from $24.1\%$ to $39.8\%$ ($17.7\%$ to $33.9\%$ on the held-out test checklists), and $4.2\%$ to $5.0\%$ more of the negative statements whose finding the checklist already listed; these pooled changes describe default-instruction behaviour and do not decompose the change in training gain. General Gemma-3 checkers of 4, 12, and 27 billion parameters differ from the training checker by less than MedGemma in point estimate, though not in an order that follows size in both populations, and 1 model per size identifies no effect of medical tuning. Under the reason-first instruction MedGemma's own format effect shifts by $-6.9\%$ and $-6.8\%$, and the differences between the checkers under both instructions include zero (Supplementary Information).

\subsection*{No comparable effect was detected for single objects in everyday photographs}
\label{sec:res_photo}

The difference could be a property of the 2 checker families rather than of the radiology checklists, so we prespecified an inference-only photograph study, frozen after its automated screen and before any checklist or checker read of a study image (Supplementary Information). Its 516 pairs are complementary VQA v2 pairs \cite{goyal2017making}, 2 COCO 2014 photographs \cite{lin2014coco} that differ in whether 1 of 20 objects is present; 2 untrained checklist models, Qwen3-VL-8B and Qwen2.5-VL-7B \cite{qwen3vl2025,bai2025qwen25vl}, record the objects they see, and the training checker's model and Gemma-3-12B read the same checklists with unmentioned objects left blank or entered as absent and, under a stated closed-world convention, as all 20 states or as the present objects alone. In its single read of the 510 pairs with all 4 checklists, the study did not detect a difference between the checkers, $-0.7\%$ $[-2.1\%, +0.7\%]$, or between the encodings, $-0.2\%$ $[-1.2\%, +0.8\%]$, so neither of its 2 prespecified effect criteria passed, while its comprehension and target-control checks for the primary checkers passed. The null's scope is narrow, since the checklist models state the target object correctly on 94.6\% and 95.4\% of images, the difference between the models is unresolved, and the checkers' model differences track a rule-based reading of the decoded states within $0.8\%$, so the radiology difference was not reproduced in a near-ceiling, inference-only comparison that differs from the radiology study in training, domain, models, and statement structure and does not identify which difference explains its absence.

\subsection*{Penalizing unmentioned findings produced complete checklists without a detected gain in discrimination}
\label{sec:res_omitcost}

The reward read every unmentioned finding as absent and never penalized it, so before any of its own training we prespecified a follow-up whose reward subtracts a quarter of the fraction of the 12 findings a checklist leaves unmentioned. All 3 of its runs list all 12 findings on every validation checklist, and paired against their matching replicated runs the training checker's rise in false alarms falls by $5.5\%$ $[+2.7\%, +8.4\%]$ while the rule-based check's rises by $3.5\%$ $[+0.1\%, +6.9\%]$, with no detected change in any checker's gain in $Y$ and the joint criterion again unmet, a pattern the held-out test checklists repeat (Supplementary Information). A complete checklist is not by itself a correct one.

\subsection*{The trained checklists carry image-specific information, concentrated in a few findings}
\label{sec:res_learned}

\begin{figure*}[t]
  \centering
  \includegraphics[width=\linewidth]{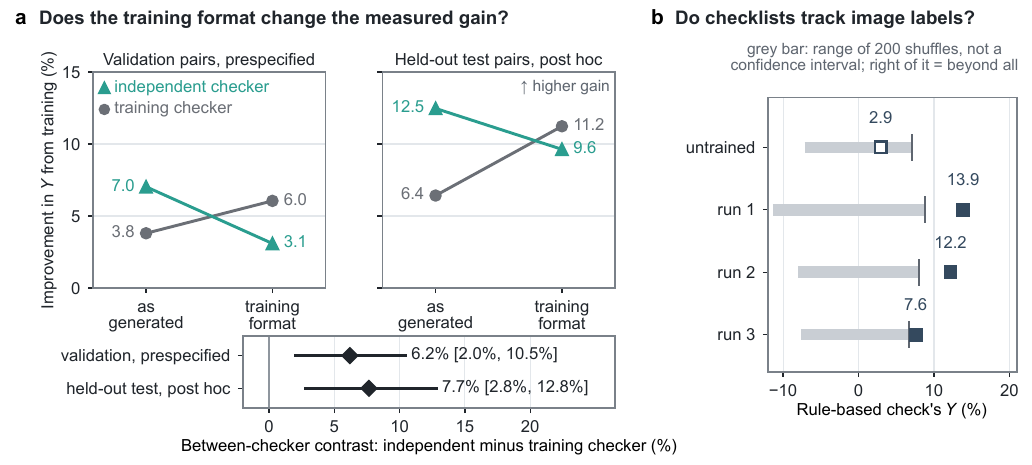}
  \caption{\textbf{What each checker measures, and whether the trained checklists describe the image.} 3 replication runs, in percent. (a) The run-averaged improvement in discrimination ($Y$) from untrained to trained checklists when the same saved checklists are read as generated and in the training format, on 237 validation pairs (prespecified) and 270 held-out test pairs (post hoc). Below, the independent checker's change from the training format to the checklists as generated minus the training checker's, with its 95\% interval over patient groups; neither checker's own change is resolved on the validation pairs. The held-out estimates differ slightly from Table~\ref{tab:main} because these rereads use the 270 pairs on which both checkers returned a verdict in every format rather than the 273 endpoint pairs. (b) Descriptive: the rule-based check's $Y$ on the 238 validation pairs for the untrained model and each run (squares) beside the full range of 200 shuffles of the checklists among images of the same finding and sentence direction (grey bars, dark tick at the maximum; not confidence intervals). A value right of the bar exceeds every shuffle.}
  \label{fig:evidence}
\end{figure*}

A gain in $Y$ could come from checklists that differ between findings without describing the particular image, so the checklists of each finding and sentence direction were shuffled among the images of that group. The rule-based check's $Y$ for the untrained checklists, $2.9\%$ on all 238 pairs, lies inside the range of its 200 shuffles, whereas no shuffle of any run reaches its real value, $13.9\%$, $12.2\%$, and $7.6\%$ against shuffle maxima of $8.8\%$, $8.0\%$, and $6.7\%$ (Fig.~\ref{fig:evidence}b), so the trained checklists carry image-specific information about the report-derived target labels. Averaged over findings and runs, the target finding's true-positive rate rises from $28.7\%$ to $40.4\%$ and its false-positive rate from $27.0\%$ to $30.8\%$, so their difference grows from $1.7\%$ to $9.7\%$ with a shift toward calling findings present.

The learning is narrow. After training, pleural effusion is marked present more often, including on images whose label is absent, and consolidation and edema are marked present less often, while the rarest findings stay almost never marked and cardiomegaly is marked on every image in 2 runs (Supplementary Information). On the held-out test pairs the gain survives removing pleural effusion's 48 pairs, $+9.2\%$ $[+3.3\%, +15.1\%]$ and $+9.8\%$ $[+2.3\%, +17.3\%]$ for the 2 checkers not used in training in a post hoc summary, though an average weighting all 8 findings equally is not resolved. The pattern follows where the reward carried signal, 42\% to 48\% of the edema and effusion groups against 2\% to 4\% for pneumothorax, lung lesion, and fracture (Supplementary Information).

\subsection*{Direct supervision and a second checklist model bound the claim}
\label{sec:res_sft}

Given the labels directly, in a single-run development comparison a supervised checklist model with rarity-weighted marks raised the rule-based check's $Y$ by $+15.1\%$ $[+7.3\%, +22.6\%]$ against $+9.7\%$ for the reinforcement run (Supplementary Information), so we claim only that a verdict on 1 sentence per image, passed through a frozen checker, carries enough of the labels to teach their discrimination.

A second checklist model, Qwen2.5-VL-7B-Instruct \cite{bai2025qwen25vl}, trained once with the unchanged recipe as a prespecified generality check, showed no detected gain, $+2.1\%$ $[-3.3\%, +7.7\%]$ under the rule-based check, and collapsed to listing 1 finding, which limits the claim to the replicated model.

\section*{Discussion}

The experiments separate 2 questions: whether the trained checklists discriminate the report-derived labels better, and whether a particular checker registers that improvement. The first was confirmed on fresh training runs and held-out patients, while the second depended on the checker and on how unmentioned findings were presented to it. The training reward never penalized unmentioned findings, and their number varied from run to run, and a model read in another format than the one it was rewarded in can be measured differently, even by the checker that trained it, which re-evaluating saved checklists reveals cheaply. Checking models range from rejecting nearly every label-consistent negative statement about an unmentioned finding to accepting nearly every one. In this task the first behaviour raises false alarms on sentences consistent with their report-derived labels, and the second misses statements that contradict the label when the unmentioned finding is labelled present. Neither behaviour estimates clinical error or alert rates, but together they show that how an evaluation reads an unmentioned finding is part of what it measures.

The same question arises wherever a report is compared with finding-level output, whether a model-written checklist, as here, or the output of a detection algorithm that reports only the findings it detects. For a radiologist, silence about a finding is not a negative finding, and structured reporting addresses this by prompting explicit statements of relevant negatives \cite{kahn2009reporting,esr2018structured,schwartz2011structured}. Automated checks need the same discipline: a stated convention for unmentioned findings, shared by the reward that trains a model and by every checker that evaluates it, and results reported under more than 1 convention when conclusions depend on it. A charge for unmentioned findings produced complete checklists without a detected improvement in discrimination, which suggests rewards that separately credit correct statements of presence and of absence, as captioning rewards separate completeness from correctness \cite{tang2026cccaption}.

The claims are deliberately narrow. Labels were extracted automatically from report text and were not adjudicated by radiologists, so every measure here is agreement with report-derived labels, not clinical accuracy, and the binary labels say nothing about the severity, laterality, or interval change a sentence may also state. The held-out patients come from the same dataset and pass the same selection rule as the development pairs, so they do not constitute external validation, and a model that sees 1 frontal image lacks the lateral views, prior examinations, and clinical context that inform a report. The trained checklist is not a stronger verdict than reading the sentence first with chain-of-thought reasoning, and what improved is discrimination of report-derived labels in a checklist that never sees the sentence, not clinical perception or faithful description of severity and location. The prespecified outcome is a trade-off, not a success: non-inferiority for false alarms was not established, which does not show that false alarms rose beyond the margin, and the learning is concentrated in a few findings. A false alarm here means rejecting a label-consistent sentence on a source image, and an incomplete checklist can make that rejection reasonable. The recipe is not shown to transfer across checklist models, the intervals describe patient variation given 3 trained models, and direct supervision had the larger single-run gain when every label was available. The split of the training format into order and entered absences, and every reread of the held-out test checklists in other formats, were declared post hoc, and the 6 further checking models were added post hoc with 1 model per size, so they cannot separate model family, size, and medical tuning. The photograph study, inference only and near ceiling, did not reproduce the difference between checkers and does not identify why.

How much improvement a checker credits to a trained image description depends on how the description states the findings it left out and on which checker reads it. The prespecified comparison of the training checker and an independent medical checker yielded a difference of 6.2\% on validation pairs, repeated post hoc at 7.7\% on held-out patients, while a photograph study with single-object statements estimated it at $-0.7\%$ $[-2.1\%, +0.7\%]$. The checklist model did learn to discriminate report-derived labels on unseen patients, with held-out gains in $Y$ of 12.6\% and 11.8\% for the 2 checkers not used in training, although the prespecified joint criterion failed on false alarms. Evaluations of trained image descriptions, in radiology and beyond, should therefore state how unmentioned findings are read and report gains across description formats and checkers.

\section*{Methods}

\subsection*{Data and matched pairs}
A checker can look accurate by accepting every sentence, so the evaluation uses 2 studies per sentence: the sentence agrees with the report-derived label of 1 and disagrees with that of the other, and the 2 studies match on the other 11 labels. Matched pairs provide them. They come from CheXpert Plus \cite{chexpertplus2024}, whose frontal studies carry report text and certain labels for 12 findings derived from the report \cite{chexpert2019}: 2 studies are paired when their labels agree on every finding but 1 and the first study's report contains a sentence about that finding, so the target finding has opposite report-derived labels in the 2 studies. Pairs are matched on labels, not on radiographic appearance. The construction is that of a concurrent study \cite{anonconcurrent2027}, unchanged, so that the question is whether a trained checklist changes the verdicts on the same evidence. The labels were extracted from the findings section of each report by the CheXpert labeler supplied with the dataset and can contain extraction errors, so agreement here means agreement with the target finding's report-derived label, not a radiologist's adjudication of the sentence or the image, and we make no claim beyond it.

Decisions made while a method is developed leak into the data they are read on, so the pairs used for development are kept apart from the pairs used for confirmation, and both from training by patient. The training pool holds 4,501 single-image prompts from 3,974 patients, and a separate 64-image set used to monitor training adds 53 further patients. Development was read on a validation pool of 507 pairs from 943 other patients, restricted to the 238 pairs whose sentence the base model accepts when given no image, since on those pairs the sentence alone does not make the model reject it on the matched image. These pairs, from 458 patients in 221 connected groups, were inspected many times while the method changed, so the development analyses on them are exploratory, apart from the prespecified learning-rate test. Confirmation uses a held-out test set of 548 pairs, up to 40 per finding and direction, from patients in neither training nor the validation pairs. Its 274 pairs, selected by the same no-image acceptance rule, were fixed before any trained checklist model was evaluated on them, and its prespecified endpoint read was performed once. Pair counts differ slightly between analyses because a pair enters an analysis only when every checker in it returned a supported or unsupported verdict on both of its images: 238 validation pairs in total, 237 in the endpoint reads, and 236 in the reads in all 4 formats; 274 held-out test pairs, 273 in the endpoint reads, and 270 in the format rereads.

\subsection*{Blinded checklist model and checking model}
To keep the sentence from shaping the description of the image, the system has 2 parts. A checklist model receives a radiograph and a fixed instruction, never the sentence, and writes a checklist prompted to cover 12 findings, each entry stating whether the finding is present, its side, an optional box, and a short note; the instruction asks it to mark a finding present only when it can name a visible sign of it. A checking model then receives the checklist and a report sentence, never the image, and decides whether the sentence is supported, so the checklist cannot depend on the sentence and any dependence of the verdict on the image has to pass through the checklist. The checklist model is Qwen3-VL-8B-Instruct \cite{qwen3vl2025} with low-rank adaptation (LoRA) \cite{lora2022}, a small set of added trainable parameters, on its language layers, with the image encoder frozen; the checking model used in training is the same base model, frozen, prompted to decide the sentence from the checklist alone. All reads use temperature 0 and request a JavaScript Object Notation (JSON) object, and the prompts as run are given in the Supplementary Information.

\subsection*{Reward and training}
A useful reward must be one that a checklist ignoring the image cannot turn into discrimination, and matched pairs provide it. During training the checklist model writes checklists for single images, the checking model decides each image's sentence from its checklist, and the checklist earns a reward of 1 when the verdict agrees with the report-derived label for that image and 0 otherwise. A checklist that is the same for every image still earns half of the reward on a pair, 1 right verdict out of 2, but it cannot separate the 2 images, and separation is what we evaluate, since the sum of the expected rewards on the 2 images of a pair, minus 1, equals the pair-level Youden index.

A reward that scores malformed checklists as wrong turned out to pay for formatting. With a strict schema in which any invalid checklist earned zero, many groups with differing rewards differed only in validity, and the checklist model learned valid, complete checklists while the findings it marked present barely changed (Supplementary Information). We therefore repair every checklist before it is checked and put it in a training format that lists the findings in a fixed order and enters every finding the model did not list as absent. This adds no finding the model marked present but asserts an absence for each unmentioned finding, a choice whose consequences the Results measure. The repairs drop entries for findings outside the 12 and repeated entries, read the strings true and false as booleans, set an invalid side to not applicable or an invalid box to null, and cut notes at 300 characters.

A verdict reward on a single sample says little without a reference for how hard the image is, so training uses group relative policy optimization (GRPO) \cite{deepseekmath2024}, a reinforcement learning method that compares several checklists sampled for the same image, in an open framework \cite{hybridflow2025} with checklists sampled through an inference server \cite{vllm2023}; each checklist's advantage is its reward relative to the other checklists sampled for the same image. A group whose checklists all earn the same reward contributes no learning signal, and in the first design more than half of the groups did, so we use dynamic sampling \cite{dapo2025}, which discards such groups and refills the batch. Each run samples 16 checklists per image at temperature 1 with up to 1,024 new tokens, uses 8 images per step, a learning rate of $10^{-4}$, LoRA rank 16 and $\alpha = 32$ on the attention and feed-forward projections, and a Kullback-Leibler penalty of 0.01 toward the initial checklist model, and trains for 40 steps on 3 48 GB A6000 graphics processing units (GPUs), 2 for the checklist model and 1 for the frozen checking model, which is never updated. The 3 replication runs differ only in their random seeds.

\subsection*{Checkers and checklist formats}
Because the training checker is also the most obvious evaluator, we never let it be the only one. Every checklist the trained model writes for evaluation is saved and read by the training checker, by MedGemma \cite{medgemma2026}, a 4-billion-parameter medical model of another family given the same checklist, sentence, and instruction (the independent checker), and by a deterministic rule-based check that uses no language model and supports the sentence exactly when the checklist's present or absent mark for the sentence's finding agrees with the sentence; it reads the checklist after the same repairs and counts an unmentioned finding as absent. The 3 checkers read the same saved checklists, so their differences are differences of reading. They answer different questions, since the rule-based check measures whether the present or absent marks discriminate the report-derived labels while the 2 checking models decide whether the checklist supports the whole sentence, so no checker is a gold standard for the others. The checking models read each checklist as generated, which is how the evaluation passes a checklist to a checker, and in the training format the training checker read during training. 2 further formats separate the training format's fixed order from its entered absences: the checklist in fixed order with unmentioned findings left out, and the checklist in the model's own order with unmentioned findings entered as absent. None of the formats changes the rule-based check.

\subsection*{Metrics}
A checker that ignores the image is still right on 1 image of every pair when it always accepts or always rejects, so the headline metric scores pairs. Writing $G$ for the fraction of pairs whose verdicts are right on both images and $B$ for the fraction wrong on both, the pair-level Youden index \cite{youden1950index} $Y = G - B$ equals the accuracy on source images plus the accuracy on matched images minus 1, that is, sensitivity plus specificity minus 1 when accepting a label-consistent sentence is the positive call, so a checker that ignores the image scores 0. Because $Y$ can rise by rejecting more sentences, we report beside it the false-alarm rate $\mathrm{FA}$, the fraction of source images whose label-consistent sentence is rejected. For each checking model, the format effect is the gain from training in $Y$ measured on checklists as generated minus the gain measured on the same checklists in the training format. The difference between the checkers subtracts the training checker's format effect from the independent checker's, pair by pair on the pairs both checkers parse, with an interval for that difference itself. Both metrics are reported in percent, and a change or a difference of either is the difference of the 2 percentages, never a ratio.

\subsection*{Reference verdicts}
References anchor the scale on the validation pairs. Under the training checker, reading the sentence first and then the image gives $Y = 10.1\%$ at $\mathrm{FA} = 37.0\%$, and $11.3\%$ at $21.0\%$ with chain-of-thought reasoning \cite{wei2022chain}; the untrained blinded checklist gives $4.6\%$ at $45.8\%$ on all 238 validation pairs; and checklists written from the labels give $68.5\%$, $53.4\%$, and $100\%$ under the training checker, the independent checker, and the rule-based check. On the held-out test pairs reading the sentence first reaches $12.4\%$ at $33.9\%$, and $16.1\%$ at $18.6\%$ with chain-of-thought reasoning, higher in point estimate than the trained checklists under the same checker, whose checklists as generated reach $10.6\%$ at $55.9\%$ on the 273 endpoint pairs (post hoc).

\subsection*{Direct supervision}
The verdict reward is computed from labels, so we also gave the checklist model those labels directly. The same model and adapter are trained by supervised learning to write the checklist its image's labels imply, in 4 variants trained once each with fixed settings (Supplementary Information), and read like the trained checklist model.

\subsection*{Prespecification and statistics}
Repeated inspection of the validation pairs could make a chance improvement look real, so evaluation plans were recorded in stages \cite{nosek2018preregistration}, and every criterion is reported whether or not it was met. The original criterion asked that a trained checklist exceed the sentence-first reference in $Y$ at a false-alarm rate of at most $42.0\%$, its own rate plus $5\%$, across runs under the training checker, and it is not changed by any later result. A prespecified learning-rate test and the development readings that followed motivated a separately prespecified replication requiring each of the 2 checkers not used in training to show a run-averaged gain in $Y$ of at least $5\%$ with the lower bound of its interval above zero and a one-sided upper bound on the false-alarm increase of at most $5\%$, on 3 fresh runs and then on the held-out test set. The replication protocol was written after its training runs had been launched and before any replicated checklist was read. The between-checker difference in the effect of switching from the checklists as generated to the training format was prespecified before any replicated validation checklist was read. Its split into the fixed order and the entered absences was declared after the first run's checklists had been read as generated and in the training format, before the 2 intermediate formats were read, and is reported as post hoc, as is every reread of the held-out test checklists in other formats. Prespecification here means that each protocol and its decision rule were written to a dated project record before the corresponding data were read, and, for the photograph study, frozen under a hash manifest; these protocol records will be released with the code, and none was deposited in a public registry. Analyses declared after related results had been read are labelled post hoc throughout, even when their own data had not yet been read.

Pairs that share a patient are not independent, so all intervals come from a paired bootstrap over connected groups of patients \cite{efron1993introduction,field2007bootstrapping}, 220 groups on the 237 validation pairs common to the endpoint reads and 245 on the 273 held-out test pairs, with 2,000 draws, as two-sided 95\% percentile intervals unless stated as one-sided bounds. Replicated results average each pair's change over the runs before resampling, so the intervals describe patient variation given the 3 trained checklist models, and contrasts recompute all quantities on the same draw. Intervals for the exploratory and post hoc comparisons are not adjusted for multiple comparisons, and the number of such intervals that exclude zero is not itself evidence.

\subsection*{Further checking models and instructions}
To ask whether the difference between the checkers reflects the independent checker's size, its medical tuning, or the instruction both checkers read, the same saved checklists were reread, in the 4 formats and in both populations, by Qwen3-VL-30B-A3B \cite{qwen3vl2025}, Phi-4-mini \cite{phi4mini2025}, OLMo-2-7B \cite{olmo2_2025}, and general Gemma-3 models of 4, 12, and 27 billion parameters \cite{gemma3_2025} with the unchanged checker messages, and by both primary checkers under 2 instructions that change only the requested reply, 1 asking for the checklist entry about the sentence's finding, quoted, or the words not listed, and 1 asking for 2 or 3 sentences of reasoning before the verdict. Every such read was declared post hoc before it was made and analysed with the code of the prespecified contrast, and a pair enters a contrast only when every read of both checkers gives a supported or unsupported verdict on both of its images. For each checker we also tabulated, on the report sentences of the same checklists, the share of verdicts accepting a sentence on the checklist as generated and with every unmentioned finding entered as absent, by whether the checklist mentions the target finding and by the sentence's polarity and agreement with its label (Supplementary Information).

\subsection*{Photograph study}
The photograph study was prespecified as an inference-only study, and its protocol, pool, labels, screen outputs, code, and pinned model revisions were frozen under a hash manifest after its automated screen and before any checklist or checker read of a study image. Its 516 image-disjoint pairs are complementary pairs of VQA v2 \cite{goyal2017making}, 2 COCO 2014 photographs \cite{lin2014coco} that differ in whether 1 of 20 object categories is present, kept when at least 8 of 10 annotators and the instance annotations agree. 2 untrained checklist models, Qwen3-VL-8B and Qwen2.5-VL-7B \cite{qwen3vl2025,bai2025qwen25vl}, record the objects they see without seeing the statement, and the training checker's model and Gemma-3-12B read the checklists with unmentioned objects left blank or entered as absent and, under a stated closed-world convention, as all 20 states or as the present objects alone. The 2 prespecified estimands are the differences between the checkers in how each rewriting moves the difference between the 2 checklist models in $Y$, each passing when its 95\% interval over 10,000 resamples of whole pairs excludes zero and its estimate is at least 5\% in size, with comprehension and target-control criteria of at least 90\% (Supplementary Information).

\subsection*{Compute}
A 40-step training run took 12.7 to 16.9 A6000 GPU-hours, and re-evaluating saved checklists by another checker or in another format took minutes on 1 GPU. Including pilots and every read up to 29 September 2026, the radiology study used 279.6 A6000, 12.3 A100, and 10.2 RTX 4090 GPU-hours on the University of Rochester computing cluster and a further 10.5 RTX PRO 6000 Blackwell and 4.1 H100 GPU-hours on the Empire AI cluster for development pilots, check runs, and data staging, the photograph study's checklist and checker jobs 4.4 A6000 GPU-hours, and the rereads by checker size and instruction 16.3 A6000 GPU-hours.

\subsection*{Code availability}
The code for checklist writing, reward computation, training configuration, the checkers, the checklist formats, and every analysis, together with the prompts and pair identifiers, is available at \url{https://github.com/ali-vosoughi/VerifyGRPO-Rad}.

\section*{Data availability}
Images, reports, and labels come from CheXpert Plus \cite{chexpertplus2024} (dataset DOI 10.71718/6nvz-pm34), available from the Stanford Center for Artificial Intelligence in Medicine and Imaging to registered users under its research use agreement, and are not redistributed by the authors; Figures 1 and 2 show 3 de-identified radiographs from the dataset as examples. The photograph study uses the COCO 2014 images and instance annotations \cite{lin2014coco} and the VQA v2 questions, answers, and complementary pairs \cite{goyal2017making}. The pair identifiers, saved checklists, verdicts, and the frozen protocol of the photograph study will be released with the code.

\section*{Acknowledgements}
We gratefully acknowledge use of the research computing resources of the Empire AI Consortium, Inc, with support from Empire State Development of the State of New York, the Simons Foundation, and the Secunda Family Foundation. Experiments also used computing resources of the Department of Computer Science, University of Rochester. This research received no specific funding.

{\small\bibliographystyle{unsrtnat}\bibliography{references}}

\begin{thebibliography}{64}
\providecommand{\natexlab}[1]{#1}
\providecommand{\url}[1]{\texttt{#1}}
\expandafter\ifx\csname urlstyle\endcsname\relax
  \providecommand{\doi}[1]{doi: #1}\else
  \providecommand{\doi}{doi: \begingroup \urlstyle{rm}\Url}\fi

\bibitem[Mahmood et~al.(2023)Mahmood, Wang, Kalra, and Yan]{mahmood2023factcheck}
Razi Mahmood, Ge~Wang, Mannudeep Kalra, and Pingkun Yan.
\newblock Fact-checking of {AI}-generated reports.
\newblock In \emph{Machine Learning in Medical Imaging (MLMI 2023), Lecture Notes in Computer Science}, pages 214--223. Springer Nature Switzerland, 2023.
\newblock \doi{10.1007/978-3-031-45676-3_22}.

\bibitem[Hardy et~al.(2024)Hardy, Kim, Ro, and Rajpurkar]{rextrust2024}
Romain Hardy, Sung~Eun Kim, Du~Hyun Ro, and Pranav Rajpurkar.
\newblock Rextrust: A model for fine-grained hallucination detection in ai-generated radiology reports, 2024.
\newblock URL \url{https://arxiv.org/abs/2412.15264}.

\bibitem[Kumar et~al.(2026)Kumar, Feng, and Tang]{mj1_2026}
Bhavesh Kumar, Dylan Feng, and Leonard Tang.
\newblock Mj1: Multimodal judgment via grounded verification, 2026.
\newblock URL \url{https://arxiv.org/abs/2603.07990}.

\bibitem[Kahn et~al.(2009)Kahn, Langlotz, Burnside, Carrino, Channin, Hovsepian, and Rubin]{kahn2009reporting}
Charles~E. Kahn, Curtis~P. Langlotz, Elizabeth~S. Burnside, John~A. Carrino, David~S. Channin, David~M. Hovsepian, and Daniel~L. Rubin.
\newblock Toward best practices in radiology reporting.
\newblock \emph{Radiology}, 252\penalty0 (3):\penalty0 852--856, 2009.
\newblock \doi{10.1148/radiol.2523081992}.

\bibitem[{European Society of Radiology}(2018)]{esr2018structured}
{European Society of Radiology}.
\newblock {ESR} paper on structured reporting in radiology.
\newblock \emph{Insights into Imaging}, 9\penalty0 (1):\penalty0 1--7, 2018.
\newblock \doi{10.1007/s13244-017-0588-8}.

\bibitem[Schwartz et~al.(2011)Schwartz, Panicek, Berk, Li, and Hricak]{schwartz2011structured}
Lawrence~H. Schwartz, David~M. Panicek, Alexandra~R. Berk, Yuelin Li, and Hedvig Hricak.
\newblock Improving communication of diagnostic radiology findings through structured reporting.
\newblock \emph{Radiology}, 260\penalty0 (1):\penalty0 174--181, 2011.
\newblock \doi{10.1148/radiol.11101913}.

\bibitem[Lambert et~al.(2024)Lambert, Morrison, Pyatkin, Huang, Ivison, Brahman, Miranda, Liu, et~al.]{lambert2024tulu}
Nathan Lambert, Jacob Morrison, Valentina Pyatkin, Shengyi Huang, Hamish Ivison, Faeze Brahman, Lester James~V. Miranda, Alisa Liu, et~al.
\newblock Tulu 3: Pushing frontiers in open language model post-training.
\newblock \emph{arXiv preprint arXiv:2411.15124}, 2024.

\bibitem[Shao et~al.(2024)Shao, Wang, Zhu, Xu, Song, Bi, Zhang, Zhang, Li, Wu, and Guo]{deepseekmath2024}
Zhihong Shao, Peiyi Wang, Qihao Zhu, Runxin Xu, Junxiao Song, Xiao Bi, Haowei Zhang, Mingchuan Zhang, Y.~K. Li, Y.~Wu, and Daya Guo.
\newblock {DeepSeekMath}: Pushing the limits of mathematical reasoning in open language models, 2024.
\newblock URL \url{https://arxiv.org/abs/2402.03300}.

\bibitem[Guo et~al.(2025)Guo, Yang, Zhang, Song, Wang, Zhu, Xu, Zhang, et~al.]{guo2025deepseek}
Daya Guo, Dejian Yang, Haowei Zhang, Junxiao Song, Peiyi Wang, Qihao Zhu, Runxin Xu, Ruoyu Zhang, et~al.
\newblock {DeepSeek-R1} incentivizes reasoning in {LLMs} through reinforcement learning.
\newblock \emph{Nature}, 645\penalty0 (8081):\penalty0 633--638, 2025.
\newblock \doi{10.1038/s41586-025-09422-z}.

\bibitem[Liu et~al.(2025)Liu, Sun, Zang, Dong, Cao, Duan, Lin, and Wang]{liu2025visualrft}
Ziyu Liu, Zeyi Sun, Yuhang Zang, Xiaoyi Dong, Yuhang Cao, Haodong Duan, Dahua Lin, and Jiaqi Wang.
\newblock Visual-{RFT}: Visual reinforcement fine-tuning.
\newblock In \emph{Proceedings of the IEEE/CVF International Conference on Computer Vision (ICCV)}, pages 2034--2044, 2025.

\bibitem[Huang et~al.(2025)Huang, Jia, Zhai, Cao, Ye, Zhao, Xu, Tang, Hu, and Lin]{huang2025visionr1}
Wenxuan Huang, Bohan Jia, Zijie Zhai, Shaosheng Cao, Zheyu Ye, Fei Zhao, Zhe Xu, Xu~Tang, Yao Hu, and Shaohui Lin.
\newblock Vision-{R1}: Incentivizing reasoning capability in multimodal large language models.
\newblock \emph{arXiv preprint arXiv:2503.06749}, 2025.

\bibitem[Shen et~al.(2025)Shen, Liu, Li, Fang, Ma, Liao, Shen, Zhang, Zhao, Zhang, Xu, and Zhao]{shen2025vlmr1}
Haozhan Shen, Peng Liu, Jingcheng Li, Chunxin Fang, Yibo Ma, Jiajia Liao, Qiaoli Shen, Zilun Zhang, Kangjia Zhao, Qianqian Zhang, Ruochen Xu, and Tiancheng Zhao.
\newblock {VLM-R1}: A stable and generalizable {R1}-style large vision-language model.
\newblock \emph{arXiv preprint arXiv:2504.07615}, 2025.

\bibitem[Pan et~al.(2025)Pan, Liu, Wu, Liu, Zhu, Li, Chen, Ouyang, and Rueckert]{pan2025medvlmr1}
Jiazhen Pan, Che Liu, Junde Wu, Fenglin Liu, Jiayuan Zhu, Hongwei~Bran Li, Chen Chen, Cheng Ouyang, and Daniel Rueckert.
\newblock {MedVLM-R1}: Incentivizing medical reasoning capability of vision-language models ({VLMs}) via reinforcement learning.
\newblock In \emph{Medical Image Computing and Computer Assisted Intervention -- MICCAI 2025}, Lecture Notes in Computer Science, pages 337--347. Springer Nature Switzerland, 2025.
\newblock \doi{10.1007/978-3-032-04981-0_32}.

\bibitem[Fisch et~al.(2020)Fisch, Lee, Chang, Clark, and Barzilay]{fisch2020capwap}
Adam Fisch, Kenton Lee, Ming-Wei Chang, Jonathan Clark, and Regina Barzilay.
\newblock {CapWAP}: Image captioning with a purpose.
\newblock In \emph{Proceedings of the 2020 Conference on Empirical Methods in Natural Language Processing (EMNLP)}, pages 8755--8768, 2020.
\newblock \doi{10.18653/v1/2020.emnlp-main.705}.

\bibitem[Xing et~al.(2025)Xing, Dong, Zang, Cao, Liang, Huang, Wang, Wu, and Lin]{caprl2025}
Long Xing, Xiaoyi Dong, Yuhang Zang, Yuhang Cao, Jianze Liang, Qidong Huang, Jiaqi Wang, Feng Wu, and Dahua Lin.
\newblock {CapRL}: Stimulating dense image caption capabilities via reinforcement learning, 2025.
\newblock URL \url{https://arxiv.org/abs/2509.22647}.
\newblock ICLR 2026.

\bibitem[Yang et~al.(2026)Yang, Xing, Dong, Zang, Cao, Wang, Zhou, Bu, Liang, Huang, Wang, Wu, and Lin]{caprlpp2026}
Penghui Yang, Long Xing, Xiaoyi Dong, Yuhang Zang, Yuhang Cao, Yibin Wang, Yujie Zhou, Jiazi Bu, Jianze Liang, Qidong Huang, Jiaqi Wang, Feng Wu, and Dahua Lin.
\newblock {CapRL++}: Unified reinforcement learning with verifiable rewards for dense image and video captioning, 2026.
\newblock URL \url{https://arxiv.org/abs/2606.09393}.

\bibitem[Tang et~al.(2026)Tang, Wang, Qi, Jiang, Hou, Zeng, and Huang]{tang2026cccaption}
Zhijiang Tang, Linhua Wang, Jiaxin Qi, Weihao Jiang, Peng Hou, Anxiang Zeng, and Jianqiang Huang.
\newblock {CCCaption}: Dual-reward reinforcement learning for complete and correct image captioning.
\newblock In \emph{Proceedings of the IEEE/CVF Conference on Computer Vision and Pattern Recognition (CVPR)}, 2026.

\bibitem[Agrawal et~al.(2018)Agrawal, Batra, Parikh, and Kembhavi]{agrawal2018dont}
Aishwarya Agrawal, Dhruv Batra, Devi Parikh, and Aniruddha Kembhavi.
\newblock Don't just assume; look and answer: Overcoming priors for visual question answering.
\newblock In \emph{Proceedings of the IEEE Conference on Computer Vision and Pattern Recognition (CVPR)}, pages 4971--4980, 2018.
\newblock \doi{10.1109/CVPR.2018.00522}.

\bibitem[Goyal et~al.(2017)Goyal, Khot, Summers-Stay, Batra, and Parikh]{goyal2017making}
Yash Goyal, Tejas Khot, Douglas Summers-Stay, Dhruv Batra, and Devi Parikh.
\newblock Making the {V} in {VQA} matter: Elevating the role of image understanding in visual question answering.
\newblock In \emph{Proceedings of the IEEE Conference on Computer Vision and Pattern Recognition (CVPR)}, pages 6325--6334, 2017.
\newblock \doi{10.1109/CVPR.2017.670}.

\bibitem[Tong et~al.(2024)Tong, Liu, Zhai, Ma, LeCun, and Xie]{tong2024eyes}
Shengbang Tong, Zhuang Liu, Yuexiang Zhai, Yi~Ma, Yann LeCun, and Saining Xie.
\newblock Eyes wide shut? exploring the visual shortcomings of multimodal {LLMs}.
\newblock In \emph{Proceedings of the IEEE/CVF Conference on Computer Vision and Pattern Recognition (CVPR)}, pages 9568--9578, 2024.

\bibitem[Chen et~al.(2024{\natexlab{a}})Chen, Li, Dong, Zhang, Zang, Chen, Duan, Wang, Qiao, Lin, and Zhao]{chen2024rightway}
Lin Chen, Jinsong Li, Xiaoyi Dong, Pan Zhang, Yuhang Zang, Zehui Chen, Haodong Duan, Jiaqi Wang, Yu~Qiao, Dahua Lin, and Feng Zhao.
\newblock Are we on the right way for evaluating large vision-language models?
\newblock In \emph{Advances in Neural Information Processing Systems}, volume~37, pages 27056--27087, 2024{\natexlab{a}}.
\newblock \doi{10.52202/079017-0850}.

\bibitem[Rohrbach et~al.(2018)Rohrbach, Hendricks, Burns, Darrell, and Saenko]{rohrbach2018object}
Anna Rohrbach, Lisa~Anne Hendricks, Kaylee Burns, Trevor Darrell, and Kate Saenko.
\newblock Object hallucination in image captioning.
\newblock In \emph{Proceedings of the 2018 Conference on Empirical Methods in Natural Language Processing}, pages 4035--4045. Association for Computational Linguistics, 2018.
\newblock \doi{10.18653/v1/D18-1437}.

\bibitem[Li et~al.(2023)Li, Du, Zhou, Wang, Zhao, and Wen]{li2023evaluating}
Yifan Li, Yifan Du, Kun Zhou, Jinpeng Wang, Xin Zhao, and Ji-Rong Wen.
\newblock Evaluating object hallucination in large vision-language models.
\newblock In \emph{Proceedings of the 2023 Conference on Empirical Methods in Natural Language Processing}, pages 292--305. Association for Computational Linguistics, 2023.
\newblock \doi{10.18653/v1/2023.emnlp-main.20}.

\bibitem[Lee et~al.(2026)Lee, Park, and Im]{mmjudgebias2026}
Sua Lee, Sanghee Park, and Jinbae Im.
\newblock Mm-judgebias: A benchmark for evaluating compositional biases in mllm-as-a-judge, 2026.
\newblock URL \url{https://arxiv.org/abs/2604.18164}.

\bibitem[nez et~al.(2026)nez, Lange, Bui, Li, Osorio, Attrach, Palakala, Kapadia, Sellami, Zhang, Zhang, and Celi]{modalens2026}
Sebasti\'an Andr\'es Cajas~Ordo\ nez, Maximin Lange, Quang Bui, Anqi~Peter Li, Felipe~Ocampo Osorio, Rafi~Al Attrach, Kushul~Reddy Palakala, Sahil Kapadia, Zakaria~Laouabdia Sellami, Xinyue Zhang, Ashley Zhang, and Leo~Anthony Celi.
\newblock Modalens: Measuring image sensitivity in report-conditioned medical vlms, 2026.
\newblock URL \url{https://arxiv.org/abs/2609.15635}.

\bibitem[Jin et~al.(2026)Jin, Zhao, Kang, Zhang, and Li]{verificationmirage2026}
Ruinan Jin, Beidi Zhao, Myeongkyun Kang, Qiong Zhang, and Xiaoxiao Li.
\newblock Verification mirage: Mapping the reliability boundary of self-verification in medical vqa, 2026.
\newblock URL \url{https://arxiv.org/abs/2605.10850}.

\bibitem[Zheng et~al.(2023)Zheng, Chiang, Sheng, Zhuang, Wu, Zhuang, Lin, Li, Li, Xing, Zhang, Gonzalez, and Stoica]{zheng2023judging}
Lianmin Zheng, Wei-Lin Chiang, Ying Sheng, Siyuan Zhuang, Zhanghao Wu, Yonghao Zhuang, Zi~Lin, Zhuohan Li, Dacheng Li, Eric Xing, Hao Zhang, Joseph Gonzalez, and Ion Stoica.
\newblock Judging {LLM}-as-a-judge with {MT-Bench} and {Chatbot Arena}.
\newblock In \emph{Advances in Neural Information Processing Systems (Datasets and Benchmarks Track)}, volume~36, pages 46595--46623, 2023.

\bibitem[Chen et~al.(2024{\natexlab{b}})Chen, Chen, Zhang, Wang, Liu, Zhou, Zhang, Wan, Zhou, and Sun]{chen2024mllmjudge}
Dongping Chen, Ruoxi Chen, Shilin Zhang, Yaochen Wang, Yinuo Liu, Huichi Zhou, Qihui Zhang, Yao Wan, Pan Zhou, and Lichao Sun.
\newblock {MLLM}-as-a-judge: Assessing multimodal {LLM}-as-a-judge with vision-language benchmark.
\newblock In \emph{Proceedings of the 41st International Conference on Machine Learning}, volume 235 of \emph{Proceedings of Machine Learning Research}, pages 6562--6595. PMLR, 2024{\natexlab{b}}.

\bibitem[Sclar et~al.(2024)Sclar, Choi, Tsvetkov, and Suhr]{sclar2024quantifying}
Melanie Sclar, Yejin Choi, Yulia Tsvetkov, and Alane Suhr.
\newblock Quantifying language models' sensitivity to spurious features in prompt design or: How {I} learned to start worrying about prompt formatting.
\newblock In \emph{International Conference on Learning Representations}, pages 25055--25083, 2024.

\bibitem[Agarwal et~al.(2026)Agarwal, Patel, Liu, Singh, Dua, Meghwani, Rowe, Avendi, Abbasi, Sheng, Ravi, and Roth]{imgtextinvariance2026}
Amit Agarwal, Hitesh~Laxmichand Patel, Meizhu Liu, Jyotika Singh, Karan Dua, Hansa Meghwani, Matthew Rowe, Michael Avendi, Yassi Abbasi, Tao Sheng, Sujith Ravi, and Dan Roth.
\newblock Do image-text metrics respect semantic invariances?, 2026.
\newblock URL \url{https://arxiv.org/abs/2605.24702}.

\bibitem[Chen et~al.(2026)Chen, Chen, Lin, and Vong]{judgeconstruct2026}
Jianlin Chen, Wenhui Chen, Ziyao Lin, and Chi~Man Vong.
\newblock A judge should know what changed: Construct validity for llm-as-a-judge evaluation, 2026.
\newblock URL \url{https://arxiv.org/abs/2608.24419}.

\bibitem[Dev et~al.(2026)Dev, Sloan, Kavner, Kong, and Sandler]{judgeharness2026}
Sunishchal Dev, Andrew Sloan, Joshua Kavner, Nicholas Kong, and Morgan Sandler.
\newblock Judge reliability harness: Stress testing the reliability of llm judges, 2026.
\newblock URL \url{https://arxiv.org/abs/2603.05399}.

\bibitem[Smit et~al.(2020)Smit, Jain, Rajpurkar, Pareek, Ng, and Lungren]{smit2020chexbert}
Akshay Smit, Saahil Jain, Pranav Rajpurkar, Anuj Pareek, Andrew Ng, and Matthew Lungren.
\newblock Combining automatic labelers and expert annotations for accurate radiology report labeling using {BERT}.
\newblock In \emph{Proceedings of the 2020 Conference on Empirical Methods in Natural Language Processing (EMNLP)}, pages 1500--1519. Association for Computational Linguistics, 2020.
\newblock \doi{10.18653/v1/2020.emnlp-main.117}.

\bibitem[Jain et~al.(2021)Jain, Agrawal, Saporta, Truong, Duong, Bui, Chambon, Zhang, Lungren, Ng, Langlotz, and Rajpurkar]{jain2021radgraph}
Saahil Jain, Ashwin Agrawal, Adriel Saporta, Steven Truong, Du~Nguyen Duong, Tan Bui, Pierre Chambon, Yuhao Zhang, Matthew Lungren, Andrew Ng, Curtis Langlotz, and Pranav Rajpurkar.
\newblock {RadGraph}: Extracting clinical entities and relations from radiology reports.
\newblock In \emph{Proceedings of the Neural Information Processing Systems Track on Datasets and Benchmarks}, volume~1, 2021.

\bibitem[Yu et~al.(2023)Yu, Endo, Krishnan, Pan, Tsai, Reis, Fonseca, Lee, Abad, Ng, Langlotz, Venugopal, and Rajpurkar]{yu2023evaluating}
Feiyang Yu, Mark Endo, Rayan Krishnan, Ian Pan, Andy Tsai, Eduardo~Pontes Reis, Eduardo Kaiser Ururahy~Nunes Fonseca, Henrique Min~Ho Lee, Zahra Shakeri~Hossein Abad, Andrew~Y. Ng, Curtis~P. Langlotz, Vasantha~Kumar Venugopal, and Pranav Rajpurkar.
\newblock Evaluating progress in automatic chest {X}-ray radiology report generation.
\newblock \emph{Patterns}, 4\penalty0 (9):\penalty0 100802, 2023.
\newblock \doi{10.1016/j.patter.2023.100802}.

\bibitem[Ostmeier et~al.(2024)Ostmeier, Xu, Chen, Varma, Blankemeier, Bluethgen, Michalson, Moseley, Langlotz, Chaudhari, and Delbrouck]{green2024}
Sophie Ostmeier, Justin Xu, Zhihong Chen, Maya Varma, Louis Blankemeier, Christian Bluethgen, Arne~Edward Michalson, Michael Moseley, Curtis Langlotz, Akshay~S Chaudhari, and Jean-Benoit Delbrouck.
\newblock Green: Generative radiology report evaluation and error notation, 2024.
\newblock URL \url{https://arxiv.org/abs/2405.03595}.
\newblock Findings of EMNLP 2024.

\bibitem[Thomas et~al.(2025)Thomas, Varma, Delbrouck, and Langlotz]{radprm2025}
Alois Thomas, Maya Varma, Jean-Benoit Delbrouck, and Curtis~P. Langlotz.
\newblock Process reward models for sentence-level verification of lvlm radiology reports, 2025.
\newblock arXiv:2510.23217.

\bibitem[Kumar(2026)]{blindrecap2026}
Yogesh Kumar.
\newblock Catching hallucinated citations in {Video-LLM} question answering: A self-verification pipeline and verifier ablation study, 2026.
\newblock URL \url{https://arxiv.org/abs/2608.15574}.

\bibitem[Gozel(2026)]{commitfirst2026}
Idil Gozel.
\newblock Commit-first {LLM} judging inherits the judge's own errors, 2026.
\newblock URL \url{https://arxiv.org/abs/2609.00088}.

\bibitem[Vosoughi et~al.(2026)]{anonconcurrent2027}
Ali Vosoughi et~al.
\newblock Verifiers look where the claim points, 2026.
\newblock Manuscript in preparation.

\bibitem[Li et~al.(2025)Li, Yu, Huang, Liang, Liu, Liu, Che, Yu, Boyd-Graber, Mi, and Yu]{visionsr1_2025}
Zongxia Li, Wenhao Yu, Chengsong Huang, Zhenwen Liang, Rui Liu, Fuxiao Liu, Jingxi Che, Dian Yu, Jordan Boyd-Graber, Haitao Mi, and Dong Yu.
\newblock Self-rewarding vision-language model via reasoning decomposition, 2025.
\newblock URL \url{https://arxiv.org/abs/2508.19652}.

\bibitem[Liu et~al.(2026)Liu, Leng, Nan, Zhu, Huang, Sun, Zhu, Chen, Wei, and Tao]{vepa2026}
Yilian Liu, Sicong Leng, Guoshun Nan, Junyi Zhu, Jiayu Huang, Minghao Sun, Xuancheng Zhu, Yisong Chen, Zexian Wei, and Xiaofeng Tao.
\newblock See first, answer later: Visual evidence pre-alignment via sufficiency-driven {RL}, 2026.
\newblock URL \url{https://arxiv.org/abs/2606.17678}.

\bibitem[Wang et~al.(2026)Wang, Xu, Wang, Xue, Peng, Chen, and Lin]{moca2026}
Haozhe Wang, Qixin Xu, Changpeng Wang, Taofeng Xue, Chong Peng, Wenhu Chen, and Fangzhen Lin.
\newblock Bad seeing or bad thinking? rewarding perception for multimodal reasoning, 2026.
\newblock URL \url{https://arxiv.org/abs/2605.14054}.
\newblock ICML 2026.

\bibitem[Amodei et~al.(2016)Amodei, Olah, Steinhardt, Christiano, Schulman, and Man{\'e}]{amodei2016concrete}
Dario Amodei, Chris Olah, Jacob Steinhardt, Paul Christiano, John Schulman, and Dan Man{\'e}.
\newblock Concrete problems in {AI} safety.
\newblock \emph{arXiv preprint arXiv:1606.06565}, 2016.

\bibitem[Skalse et~al.(2022)Skalse, Howe, Krasheninnikov, and Krueger]{skalse2022defining}
Joar Skalse, Nikolaus Howe, Dmitrii Krasheninnikov, and David Krueger.
\newblock Defining and characterizing reward gaming.
\newblock In \emph{Advances in Neural Information Processing Systems}, volume~35, pages 9460--9471, 2022.
\newblock \doi{10.52202/068431-0687}.

\bibitem[Gao et~al.(2023)Gao, Schulman, and Hilton]{gao2023scaling}
Leo Gao, John Schulman, and Jacob Hilton.
\newblock Scaling laws for reward model overoptimization.
\newblock In \emph{Proceedings of the 40th International Conference on Machine Learning}, volume 202 of \emph{Proceedings of Machine Learning Research}, pages 10835--10866. PMLR, 2023.

\bibitem[{Gemma Team}(2025)]{gemma3_2025}
{Gemma Team}.
\newblock Gemma 3 technical report, 2025.

\bibitem[Lin et~al.(2014)Lin, Maire, Belongie, Hays, Perona, Ramanan, Doll{\'a}r, and Zitnick]{lin2014coco}
Tsung-Yi Lin, Michael Maire, Serge Belongie, James Hays, Pietro Perona, Deva Ramanan, Piotr Doll{\'a}r, and C.~Lawrence Zitnick.
\newblock Microsoft {COCO}: Common objects in context.
\newblock In \emph{European Conference on Computer Vision (ECCV)}, pages 740--755, 2014.
\newblock \doi{10.1007/978-3-319-10602-1_48}.

\bibitem[Bai et~al.(2025{\natexlab{a}})Bai, Cai, Chen, Chen, Chen, Cheng, Deng, Ding, Gao, Ge, Ge, Guo, Huang, Huang, Huang, Hui, Lin, Liu, Liu, Men, Ren, Song, Tang, Wang, Xie, Xu, Yang, Yu, Zhang, Zheng, and Zhou]{qwen3vl2025}
Shuai Bai, Yuxuan Cai, Ruizhe Chen, Keqin Chen, Xionghui Chen, Zesen Cheng, Lianghao Deng, Wei Ding, Chang Gao, Chunjiang Ge, Wenbin Ge, Zhifang Guo, Qidong Huang, Jie Huang, Fei Huang, Binyuan Hui, Junyang Lin, Xuejing Liu, Dayiheng Liu, Rui Men, Xuancheng Ren, Sibo Song, Jun Tang, Peng Wang, Tianbao Xie, Haiyang Xu, An~Yang, Bowen Yu, Hang Zhang, Bo~Zheng, and Jingren Zhou.
\newblock Qwen3-vl technical report, 2025{\natexlab{a}}.
\newblock URL \url{https://arxiv.org/abs/2511.21631}.

\bibitem[Bai et~al.(2025{\natexlab{b}})Bai, Chen, Liu, Wang, Ge, Song, Dang, Wang, et~al.]{bai2025qwen25vl}
Shuai Bai, Keqin Chen, Xuejing Liu, Jialin Wang, Wenbin Ge, Sibo Song, Kai Dang, Peng Wang, et~al.
\newblock {Qwen2.5-VL} technical report.
\newblock \emph{arXiv preprint arXiv:2502.13923}, 2025{\natexlab{b}}.

\bibitem[Chambon et~al.(2024)Chambon, Delbrouck, Sounack, Huang, Chen, Varma, Truong, Chuong, and Langlotz]{chexpertplus2024}
Pierre Chambon, Jean-Benoit Delbrouck, Thomas Sounack, Shih-Cheng Huang, Zhihong Chen, Maya Varma, Steven~QH Truong, Chu~The Chuong, and Curtis~P. Langlotz.
\newblock Chexpert plus: Augmenting a large chest x-ray dataset with text radiology reports, patient demographics and additional image formats, 2024.
\newblock URL \url{https://arxiv.org/abs/2405.19538}.
\newblock Dataset DOI: 10.71718/6nvz-pm34.

\bibitem[Irvin et~al.(2019)Irvin, Rajpurkar, Ko, Yu, Ciurea-Ilcus, Chute, Marklund, Haghgoo, Ball, Shpanskaya, Seekins, Mong, Halabi, Sandberg, Jones, Larson, Langlotz, Patel, Lungren, and Ng]{chexpert2019}
Jeremy Irvin, Pranav Rajpurkar, Michael Ko, Yifan Yu, Silviana Ciurea-Ilcus, Chris Chute, Henrik Marklund, Behzad Haghgoo, Robyn Ball, Katie Shpanskaya, Jayne Seekins, David~A. Mong, Safwan~S. Halabi, Jesse~K. Sandberg, Ricky Jones, David~B. Larson, Curtis~P. Langlotz, Bhavik~N. Patel, Matthew~P. Lungren, and Andrew~Y. Ng.
\newblock {CheXpert}: A large chest radiograph dataset with uncertainty labels and expert comparison.
\newblock In \emph{Proceedings of the AAAI Conference on Artificial Intelligence}, volume~33, pages 590--597, 2019.
\newblock \doi{10.1609/aaai.v33i01.3301590}.
\newblock URL \url{https://ojs.aaai.org/index.php/AAAI/article/view/3834}.

\bibitem[Hu et~al.(2022)Hu, Shen, Wallis, Allen-Zhu, Li, Wang, Wang, and Chen]{lora2022}
Edward~J. Hu, Yelong Shen, Phillip Wallis, Zeyuan Allen-Zhu, Yuanzhi Li, Shean Wang, Lu~Wang, and Weizhu Chen.
\newblock {LoRA}: Low-rank adaptation of large language models.
\newblock In \emph{International Conference on Learning Representations}, 2022.
\newblock URL \url{https://openreview.net/forum?id=nZeVKeeFYf9}.

\bibitem[Sheng et~al.(2025)Sheng, Zhang, Ye, Wu, Zhang, Zhang, Peng, Lin, and Wu]{hybridflow2025}
Guangming Sheng, Chi Zhang, Zilingfeng Ye, Xibin Wu, Wang Zhang, Ru~Zhang, Yanghua Peng, Haibin Lin, and Chuan Wu.
\newblock {HybridFlow}: A flexible and efficient {RLHF} framework.
\newblock In \emph{Proceedings of the Twentieth European Conference on Computer Systems (EuroSys '25)}, pages 1279--1297. ACM, 2025.
\newblock \doi{10.1145/3689031.3696075}.
\newblock URL \url{https://arxiv.org/abs/2409.19256}.

\bibitem[Kwon et~al.(2023)Kwon, Li, Zhuang, Sheng, Zheng, Yu, Gonzalez, Zhang, and Stoica]{vllm2023}
Woosuk Kwon, Zhuohan Li, Siyuan Zhuang, Ying Sheng, Lianmin Zheng, Cody~Hao Yu, Joseph~E. Gonzalez, Hao Zhang, and Ion Stoica.
\newblock Efficient memory management for large language model serving with pagedattention, 2023.
\newblock URL \url{https://arxiv.org/abs/2309.06180}.
\newblock SOSP 2023.

\bibitem[Yu et~al.(2025)Yu, Zhang, Zhu, Yuan, Zuo, Yue, Dai, Fan, Liu, Liu, et~al.]{dapo2025}
Qiying Yu, Zheng Zhang, Ruofei Zhu, Yufeng Yuan, Xiaochen Zuo, Yu~Yue, Weinan Dai, Tiantian Fan, Gaohong Liu, Lingjun Liu, et~al.
\newblock {DAPO}: An open-source {LLM} reinforcement learning system at scale, 2025.
\newblock URL \url{https://arxiv.org/abs/2503.14476}.
\newblock NeurIPS 2025.

\bibitem[Sellergren et~al.(2025)Sellergren, Kazemzadeh, Jaroensri, Kiraly, Traverse, Kohlberger, Xu, Jamil, Hughes, Lau, Chen, Mahvar, Yatziv, Azizi, Liu, Shetty, Steiner, Golden, and Yang]{medgemma2026}
Andrew Sellergren, Sahar Kazemzadeh, Tiam Jaroensri, Atilla Kiraly, Madeleine Traverse, Timo Kohlberger, Shawn Xu, Fayaz Jamil, C\'ian Hughes, Charles Lau, Justin Chen, Fereshteh Mahvar, Liron Yatziv, Shekoofeh Azizi, Yun Liu, Shravya Shetty, David~F. Steiner, Daniel Golden, and Lin Yang.
\newblock Medgemma technical report, 2025.
\newblock URL \url{https://arxiv.org/abs/2507.05201}.

\bibitem[Youden(1950)]{youden1950index}
W.~J. Youden.
\newblock Index for rating diagnostic tests.
\newblock \emph{Cancer}, 3\penalty0 (1):\penalty0 32--35, 1950.
\newblock \doi{10.1002/1097-0142(1950)3:1<32::AID-CNCR2820030106>3.0.CO;2-3}.

\bibitem[Wei et~al.(2022)Wei, Wang, Schuurmans, Bosma, Ichter, Xia, Chi, Le, and Zhou]{wei2022chain}
Jason Wei, Xuezhi Wang, Dale Schuurmans, Maarten Bosma, Brian Ichter, Fei Xia, Ed~Chi, Quoc~V. Le, and Denny Zhou.
\newblock Chain-of-thought prompting elicits reasoning in large language models.
\newblock In \emph{Advances in Neural Information Processing Systems}, volume~35, pages 24824--24837, 2022.
\newblock \doi{10.52202/068431-1800}.

\bibitem[Nosek et~al.(2018)Nosek, Ebersole, DeHaven, and Mellor]{nosek2018preregistration}
Brian~A. Nosek, Charles~R. Ebersole, Alexander~C. DeHaven, and David~T. Mellor.
\newblock The preregistration revolution.
\newblock \emph{Proceedings of the National Academy of Sciences}, 115\penalty0 (11):\penalty0 2600--2606, 2018.
\newblock \doi{10.1073/pnas.1708274114}.

\bibitem[Efron and Tibshirani(1993)]{efron1993introduction}
Bradley Efron and Robert~J. Tibshirani.
\newblock \emph{An Introduction to the Bootstrap}, volume~57 of \emph{Monographs on Statistics and Applied Probability}.
\newblock Chapman \& Hall, New York, 1993.

\bibitem[Field and Welsh(2007)]{field2007bootstrapping}
C.~A. Field and A.~H. Welsh.
\newblock Bootstrapping clustered data.
\newblock \emph{Journal of the Royal Statistical Society Series B: Statistical Methodology}, 69\penalty0 (3):\penalty0 369--390, 2007.
\newblock \doi{10.1111/j.1467-9868.2007.00593.x}.

\bibitem[Abouelenin et~al.(2025)Abouelenin, Ashfaq, Atkinson, et~al.]{phi4mini2025}
Abdelrahman Abouelenin, Atabak Ashfaq, Adam Atkinson, et~al.
\newblock Phi-4-mini technical report: Compact yet powerful multimodal language models via mixture-of-loras, 2025.
\newblock URL \url{https://arxiv.org/abs/2503.01743}.

\bibitem[{Team OLMo} et~al.(2025){Team OLMo}, Walsh, Soldaini, et~al.]{olmo2_2025}
{Team OLMo}, Pete Walsh, Luca Soldaini, et~al.
\newblock 2 {OLMo} 2 furious, 2025.
\newblock URL \url{https://arxiv.org/abs/2501.00656}.

\end{thebibliography}

\clearpage
\appendix
\setcounter{section}{0}
\setcounter{figure}{0}
\setcounter{table}{0}
\renewcommand{\thesection}{S\arabic{section}}
\renewcommand{\thefigure}{S\arabic{figure}}
\renewcommand{\thetable}{S\arabic{table}}
\renewcommand{\theHsection}{S\arabic{section}}
\renewcommand{\theHfigure}{S\arabic{figure}}
\renewcommand{\theHtable}{S\arabic{table}}
\renewcommand{\figurename}{Figure}
\begin{center}
{\Large\bfseries Supplementary Information\par}
\vspace{0.5em}
{\large Unmentioned Checklist Findings Change How Reinforcement Learning Appears to Improve Chest Radiograph Report Checking\par}
\vspace{0.8em}
{Ali Vosoughi, Akhil Kasturi, Chenliang Xu, Axel Wismueller\par}
\end{center}
\noindent\textbf{Terms.} The main text uses clinical terms for the technical terms kept in this supplement: the writer is the checklist model, a record is its findings checklist, a judge is a checking model (the training judge is the training checker and MedGemma the independent checker), the rule-based or flag reader is the rule-based check, an omitted finding is an unmentioned finding, the original form is the checklist as generated, the reward-scoring (canonical) form is the training format, the false-strike rate FS is the false-alarm rate FA, and registered means prespecified. Every number is the same in both documents.

\section{Development analyses}
\label{sec:supp_discovery}

This section documents development on the validation population, whose 238 pairs, selected by the no-image acceptance rule, were inspected many times while the method changed, the registered learning-rate test, and diagnostics computed on the training images. Throughout, $Y$ is the pair-level Youden index, the fraction of pairs judged right on both images minus the fraction judged wrong on both, and FS the false-strike rate, the fraction of report images whose true sentence is rejected. It explains how the recipe of the registered replication was found and what first suggested the analyses that were then registered. None of it is confirmation, and analyses computed after their estimates were seen are marked post hoc.

\subsection{A reward that punishes malformed records teaches formatting}

The first design scored any record that failed a strict schema as wrong, at a learning rate of $10^{-5}$ with 8 records per image. Over 70 steps no change in the verdicts on the validation pairs was detected, with $Y$ between 2.9\% and 6.7\% and the false-strike rate between 43.7\% and 47.5\% at every saved step, and a paired change from start to end of $-1.3\%$ $[-5.0\%, +2.5\%]$. The reward decomposition suggests why. Sampling 8 records per image at the training temperature on 64 held-out images from the training split, 17\% of groups differed only because some records were invalid, and those groups resolved as the writer learned the schema, with invalid samples falling from 8.4\% to 0.4\% while groups whose valid records disagreed on content shrank from 28\% to 22\%. The writer came to list all 12 findings, up from 10.4 on average, but the number of findings it marked present stayed at 4.6 to 4.9. Its single-sample success on those images rose from 44\% to 47\% while its best-of-8 success stayed at 59\%, the signature of a policy that sharpens what it already samples.

\subsection{Content alone, and the learning rate that moves it}

Normalizing records before judging them, so that a formatting error no longer costs the reward of a usable record, and adding dynamic sampling with wider exploration, did not change the verdicts detectably at the original learning rate either, with a paired change in $Y$ of $+0.4\%$ $[-4.1\%, +4.9\%]$ after 60 steps. The untrained writer could in principle earn much more reward, since for 72\% of the same held-out images at least 1 of 64 sampled records leads the judge to the right verdict, although 11 of those images share a patient with the training set. A bounded learning-rate test was therefore registered, comparing learning rates of $3\times10^{-5}$ and $10^{-4}$ on a fixed set of 64 training images and extending the better setting to 40 steps. By its registered criteria it failed. Probe accuracy rose from 45.9\% to 51.6\%, short of the required rise of 10\%, and on the validation pairs the training judge's $Y$ rose by $4.2\%$ $[-1.7\%, +10.6\%]$, from 4.6\% to 8.8\%, while its false-strike rate rose from 45.8\% to 53.8\%, above the criterion's limit. That verdict stands as registered. The registered replication of the main paper repeats the recipe of the 40-step setting unchanged.

\subsection{The development run under 3 readers}

\begin{table}[h]
  \centering\small
  \begin{tabular}{lcccccc}
    \toprule
    & \multicolumn{2}{c}{Training judge} & \multicolumn{2}{c}{Independent judge} & \multicolumn{2}{c}{Rule-based reader} \\
    Record & $Y$ & FS & $Y$ & FS & $Y$ & FS \\
    \midrule
    Untrained writer & 4.6 & 45.8 & 2.1 & 36.3 & 2.9 & 45.0 \\
    Development run, step 40 & 8.8 & 53.8 & 12.2 & 35.9 & 12.6 & 44.1 \\
    Change & $+$4.2 & $+$8.0 & $+$10.1 & $-$0.4 & $+$9.7 & $-$0.8 \\
    \midrule
    Labels as flags & 68.5 & 27.7 & 53.4 & 4.2 & 100.0 & 0.0 \\
    \bottomrule
  \end{tabular}
  \caption{The development run's records read in their original form by 3 readers on the 238 validation pairs, all entries in percent (237 common pairs for the independent judge's paired untrained-to-trained comparison, since its output for 1 trained record contained no parseable verdict, while the label row uses all 238). $Y$ is the pair-level Youden index and FS the false-strike rate. Changes are paired, with 95\% intervals of $[-1.7\%, +10.6\%]$, $[+2.1\%, +18.1\%]$, and $[+3.3\%, +16.5\%]$ for $Y$ and $[+4.1\%, +12.1\%]$, $[-5.6\%, +4.8\%]$, and $[-5.5\%, +3.8\%]$ for FS. Records written from the labels show that each reader has room to register a perfect flag record, and that the judges fall short of it even then, as they also assess sentence details that the flags do not specify.}
  \label{tab:supp_discovery_readers}
\end{table}

The readers that did not train the writer showed a gain of about 10\% on the development run, and the training judge less, with more false strikes (Table~\ref{tab:supp_discovery_readers}). Contrasted post hoc on the same patient resamples, the rule-based reader's gain exceeds the training judge's by $5.5\%$ $[+0.4\%, +10.8\%]$, and the training judge's rise in false strikes exceeds the rule-based reader's by $8.9\%$ $[+4.3\%, +13.5\%]$ and the independent judge's by $8.4\%$ $[+3.8\%, +13.1\%]$. These contrasts were then declared for the replication, where the 2 false-strike contrasts repeated and the gain contrast did not resolve. On the development run, without the 69 pleural effusion pairs the rule-based reader's gain was $+6.0\%$ and the independent judge's $+8.9\%$, and with findings weighted equally $+8.6\%$ and $+9.4\%$; these are point estimates, because the intervals of this development summary could not be recomputed.

\subsection{The record form on the development run}

\begin{table}[h]
  \centering\small
  \setlength{\tabcolsep}{4pt}
  \begin{tabular}{lcccc}
    \toprule
    & \multicolumn{2}{c}{Training judge} & \multicolumn{2}{c}{Independent judge} \\
    Records read & $\Delta Y$ & $\Delta$FS & $\Delta Y$ & $\Delta$FS \\
    \midrule
    Original form & $+$4.2 {\scriptsize[$-$1.7, $+$10.6]} & $+$8.0 {\scriptsize[$+$4.1, $+$12.1]} & $+$10.1 {\scriptsize[$+$2.1, $+$18.1]} & $-$0.4 {\scriptsize[$-$5.6, $+$4.8]} \\
    Original form, sides not applicable & $+$3.8 {\scriptsize[$-$2.4, $+$10.2]} & $+$7.6 {\scriptsize[$+$3.7, $+$11.5]} & $+$11.4 {\scriptsize[$+$3.4, $+$19.2]} & $-$0.4 {\scriptsize[$-$5.5, $+$4.6]} \\
    \midrule
    Reward-scoring form, every field & $+$6.7 {\scriptsize[$+$0.4, $+$13.2]} & $+$6.3 {\scriptsize[$+$2.1, $+$10.4]} & $+$6.3 {\scriptsize[$-$1.3, $+$13.6]} & $-$0.4 {\scriptsize[$-$5.0, $+$4.3]} \\
    Reward-scoring form, notes removed & $+$7.6 {\scriptsize[$+$1.7, $+$13.6]} & $+$3.4 {\scriptsize[$-$0.8, $+$7.6]} & $+$5.9 {\scriptsize[$-$1.7, $+$13.7]} & $-$2.5 {\scriptsize[$-$8.0, $+$2.9]} \\
    Reward-scoring form, flags only & $+$8.0 {\scriptsize[$+$1.7, $+$14.5]} & $+$2.9 {\scriptsize[$-$1.7, $+$7.5]} & $+$3.4 {\scriptsize[$-$3.0, $+$10.1]} & $+$3.4 {\scriptsize[$-$1.7, $+$8.2]} \\
    \bottomrule
  \end{tabular}
  \caption{The development run's saved records read in different forms on the 238 validation pairs, all entries in percent (237 for the independent judge), each form applied to both the untrained and the trained records. Entries are paired changes in the pair-level Youden index ($\Delta Y$) and the false-strike rate ($\Delta$FS) with 95\% intervals over patient groups.}
  \label{tab:supp_discovery_form}
\end{table}

The first ablations removed the notes and reduced records to flags, and both raised the training judge's gain (Table~\ref{tab:supp_discovery_form}), which at first suggested that the notes hid the learning from it. That reading was confounded by the change of form. Both ablations start from the reward-scoring form, whereas the full row is the original record, so they changed the form as well as the content. A condition that only changes the form separates the 2. In the reward-scoring form both judges register about the same gain, $+6.7\%$ and $+6.3\%$, the difference between the judges' responses to the change of form is $6.3\%$ $[+1.3\%, +11.2\%]$ post hoc, and within the reward-scoring form the notes change neither judge's gain detectably, by $-0.8\%$ $[-3.9\%, +2.1\%]$ and $+0.4\%$ $[-6.0\%, +6.8\%]$. Separating the form into its order and its imputed absences pointed to the absences, $4.4\%$ $[0.0\%, +8.9\%]$ against $1.9\%$ $[-0.4\%, +4.3\%]$ for the order. A crossed variant that pairs the untrained writer's flags with the trained notes and the trained flags with the untrained writer's notes agrees, since the trained flags raise the training judge's $Y$ by about $6.5\%$ whichever notes accompany them. This analysis, with the between-judge contrast of the form as its primary quantity, was then registered for the replicated records, where it repeated at $6.2\%$ $[+2.0\%, +10.5\%]$.

The notes did add false strikes for the training judge, and removing them from both the untrained and the trained records in the reward-scoring form reduces its rise from $6.3\%$ to $3.4\%$, a difference of $3.0\%$ $[+0.8\%, +5.5\%]$. Its explanations name what the notes add, a severity, a location, or a co-mentioned finding that the sentence does not share, or a note that contradicts a hedge, as when a record describes a heart wider than half the thorax and the sentence reports minimal cardiomegaly. Among true sentences whose target flag the trained record gets right but which the judge still rejects, a count that rises from 11 to 27, 14 are cases whose flag training corrected.

Splitting the pairs by the lexical rule of Sec.~\ref{sec:supp_split}, fixed before the split was computed, 94 report sentences state a severity, location, or change and 144 do not. On the detailed sentences the training judge strikes 83\% of true sentences even from the untrained records, and its whole gain on the records in the original form falls on the plain sentences, $+6.9\%$ against $0.0\%$, whereas the rule-based reader has nearly identical point gains on the 2, $+9.7\%$ and $+9.6\%$. The split is confounded with the finding, since atelectasis and lung lesion sentences are almost all detailed and pleural effusion and pneumothorax sentences mostly plain, so we read it only as locating the effect. Whether a judge should reject a sentence whose details the record does not state is a separate question that no ablation here settles.

\subsection{Where the reward carried signal}

\begin{table}[h]
  \centering\small
  \begin{tabular}{lcc}
    \toprule
    & Development run & First registered run \\
    Finding & images / groups / informative & images / groups / informative \\
    \midrule
    Pleural effusion & 252 / 265 / 118 (45) & 256 / 273 / 124 (45) \\
    Edema & 229 / 229 / 110 (48) & 283 / 283 / 120 (42) \\
    Consolidation & 200 / 211 / 68 (32) & 200 / 213 / 61 (29) \\
    Cardiomegaly & 140 / 140 / 46 (33) & 158 / 158 / 24 (15) \\
    Atelectasis & 132 / 132 / 21 (16) & 167 / 167 / 30 (18) \\
    Pneumothorax & 252 / 289 / 7 (2) & 290 / 334 / 13 (4) \\
    Lung lesion & 132 / 132 / 5 (4) & 118 / 118 / 2 (2) \\
    Fracture & 126 / 126 / 2 (2) & 143 / 143 / 3 (2) \\
    \bottomrule
  \end{tabular}
  \caption{Training exposure per finding, from the per-sentence reward logs of the training images, shares in parentheses in percent. A group is a single image's sentence over its 16 records, and a group is informative when its mean reward lies strictly between 0 and 1, the only groups that carry a verdict-reward signal.}
  \label{tab:supp_exposure}
\end{table}

Table~\ref{tab:supp_exposure} counts the groups that carried a verdict-reward signal. Pneumothorax was scored most often and almost never carried one, whereas pleural effusion and edema produced one in close to half of their groups, which matches the findings that were learned.

\section{Prompts as run}
\label{sec:supp_prompts}

The writer receives a system message and 1 user message whose first element is the image. The judge receives a system message and 1 user message holding the serialized record and the sentence, never the image. The independent judge receives exactly the judge's messages. All reads use temperature 0 and request a JavaScript Object Notation (JSON) object.

\subsection{Writer, system message}
\begin{Verbatim}[fontsize=\footnotesize,breaklines]
You are reading a chest radiograph and recording what you see. Answer only with the JSON object asked for. Do not add commentary.
\end{Verbatim}
\subsection{Writer, user message, after the image}
\begin{Verbatim}[fontsize=\footnotesize,breaklines]
A chest radiograph is attached. Record what it shows, finding by finding, before any claim about it is made. A finding may be marked present ONLY if you can name a specific sign of it that is visible in this image, for example blunting of a costophrenic angle, a visible pleural line with no lung markings beyond it, a cardiac silhouette wider than half the thorax, or an air bronchogram inside an opacity. Write that sign in the note. If you cannot name such a sign for this image, the finding is absent. When in doubt, absent. Most radiographs show none or one of the findings listed.
Findings to cover: Enlarged Cardiomediastinum, Cardiomegaly, Lung Opacity, Lung Lesion, Edema, Consolidation, Pneumonia, Atelectasis, Pneumothorax, Pleural Effusion, Pleural Other, Fracture.
Reply as {"record": [{"finding": "<name>", "present": true or false, "side": "left" or "right" or "bilateral" or "not applicable", "box": [x1, y1, x2, y2] or null, "note": "<short>"}], "other": "<anything notable not in the list, one sentence>"}
\end{Verbatim}
\subsection{Judge, system message}
\begin{Verbatim}[fontsize=\footnotesize,breaklines]
You are verifying a single sentence taken from a chest radiograph report. Sentences are sometimes true of the radiograph and sometimes not; both verdicts are common. Answer only with the JSON object asked for. Do not add commentary.
\end{Verbatim}
\subsection{Judge, user message}
\begin{Verbatim}[fontsize=\footnotesize,breaklines]
Here is a record of what a chest radiograph shows, written before any claim was made:
{evidence}

Sentence from the report of that radiograph:
{claim}

Using only the record, judge whether the sentence is true.
Reply as {"verdict": "supported" or "unsupported", "reason": "<one short sentence>"}
\end{Verbatim}

\section{Preparing records for reward scoring}
\label{sec:supp_canon}

Before a training record is judged it is put in a reward-scoring form, which adds no finding the writer marked present but imputes absence for every finding the writer left out. Entries whose finding is not among the 12, and repeated entries of a finding after the first, are dropped. A presence value that is a boolean is kept, the strings true and false are read as booleans, and any other value drops the entry. A side outside the 4 allowed values becomes not applicable, a box that is not 4 finite numbers becomes null, a note that is not a string becomes empty, and a note longer than 300 characters is cut. Findings the writer did not list are recorded as absent with no side, box, or note, an assertion the writer did not make, whose effect on each judge the main paper measures. A record that already satisfies the strict schema receives exactly the serialization the strict design gave it, so its reward is unchanged, and only output containing no record object is rejected for format. During training the judge reads the normalized entry list in compact JSON. The rule-based reader and the transformed reading forms share these repairs, while the original form passes the writer's record unchanged, and the order of the entries and the handling of omitted findings depend on the form, as described below.

\section{Training and reading configuration}
\label{sec:supp_config}

\begin{table}[h]
  \centering\small
  \begin{tabular}{lp{0.72\linewidth}}
    \toprule
    Item & Setting \\
    \midrule
    Writer and judge & Qwen3-VL-8B-Instruct, bfloat16; judge frozen and served separately \\
    Adapter & low-rank adaptation (LoRA) rank 16, $\alpha = 32$, dropout 0, on q, k, v, o, gate, up, down projections of the language model; vision tower untouched \\
    Algorithm & group relative policy optimization, Kullback-Leibler (KL) loss to the initial writer (coefficient 0.01, low-variance estimator), no entropy bonus \\
    Dynamic sampling & groups with no reward variation discarded and the batch refilled \\
    Rollouts & 16 records per image, temperature 1.0, up to 1,024 new tokens \\
    Batch & 8 images per step, mini-batch 4, learning rate $10^{-4}$, 40 steps, adapters saved every 10 steps \\
    Seeds & 101, 202, 303 (runs 1, 2, 3) in the actor, reference, rollout, and data-loader seeds and in the data order \\
    Software & verl 0.9.1 (fully sharded data parallel, FSDP2), vLLM 0.24.0, PyTorch 2.11, Transformers 5.9, parameter-efficient fine-tuning library (PEFT) 0.21 \\
    Hardware & 3 48 GB A6000 per run (2 for the policy with tensor-parallel rollout, 1 for the judge) \\
    Reading & a separate environment with vLLM 0.29.0, PyTorch 2.13, and Transformers 5.17, temperature 0, 1 48 GB A6000 (or 1 80 GB A100 or 24 GB RTX 4090) per read \\
    Independent judge & MedGemma-4B-it, same messages, temperature 0 \\
    \bottomrule
  \end{tabular}
  \caption{Configuration of the registered replication, identical to the learning-rate run except for the seeds.}
\end{table}

\section{Training dynamics}
\label{sec:supp_dynamics}

\begin{figure*}[t]
  \centering
  \includegraphics[width=\textwidth]{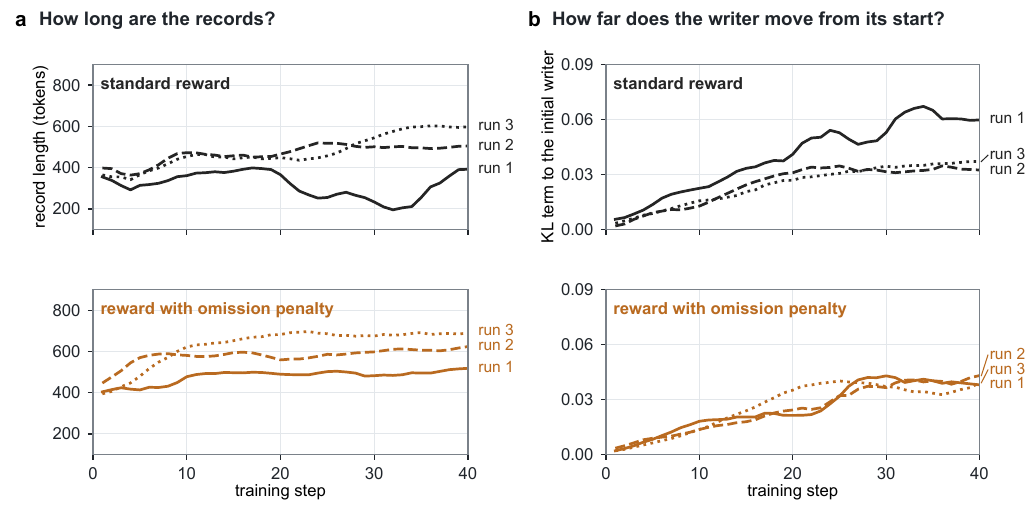}
  \caption{Training logs of the 3 replication runs with the standard reward (black) and the 3 runs with the omission penalty (ochre), runs as line styles, each curve a centred 5-step running mean of the logged steps. (a) Mean record length in tokens. (b) The Kullback-Leibler divergence term to the initial writer in the training loss.}
  \label{fig:supp_training}
\end{figure*}

The reward logged during training cannot show what the writer learned, because the framework computes it on the training batch after dynamic sampling has replaced every group whose 16 records earned the same reward, so it is conditioned on groups with reward variation and is not the judge's unfiltered accuracy. Its observed means over each run's 40 steps lie between 0.50 and 0.54, and the evaluation reads of the main paper remain the measure of what was learned. The record length is the clearest trace in the logs (Fig.~\ref{fig:supp_training}a). Runs 2 and 3 of the standard reward lengthen their records, from 422 and 411 tokens at the first step to means of 498 and 597 over the last 5 steps, while run 1 shortens them to 167 tokens at step 33 before they lengthen again to 414 at step 40, and this run also moves furthest from the initial writer, with a divergence term of 0.060 over its last 5 steps against 0.033 and 0.037. Under the omission cost every run lengthens its records, to means of 511, 608, and 687 tokens over the last 5 steps, consistent with the pressure to list all 12 findings, and these runs also used more compute than the registered ones (Sec.~\ref{sec:supp_compute}).

\section{Endpoints per run}
\label{sec:supp_perseed}

\begin{figure*}[t]
  \centering
  \includegraphics[width=\textwidth]{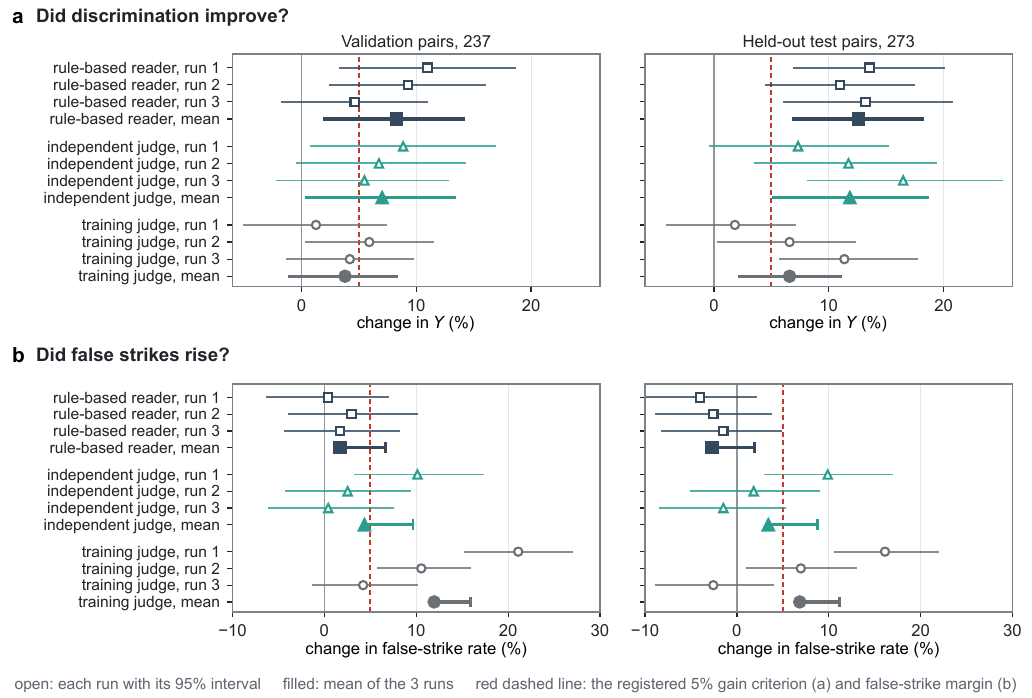}
  \caption{Registered endpoints for each training run and their mean, records read in their original form, in percent. (a) Change in the pair-level Youden index $Y$ from the untrained writer, with 95\% intervals over patient groups. (b) Change in the false-strike rate, each run with its 95\% interval and the mean with its one-sided 95\% upper bound. The red dashed line marks the registered $5\%$ gain criterion in (a) and the $5\%$ false-strike margin in (b), and the 237 validation and 273 held-out test pairs are those common to all endpoint reads.}
  \label{fig:supp_perrun}
\end{figure*}

\begin{table}[h]
  \centering\footnotesize
  \setlength{\tabcolsep}{4pt}
  \begin{tabular}{llcccc}
    \toprule
    & & \multicolumn{2}{c}{Validation, 237 pairs} & \multicolumn{2}{c}{Held-out test, 273 pairs} \\
    Reader & Writer & $\Delta Y$ & $\Delta$FS & $\Delta Y$ & $\Delta$FS \\
    \midrule
    Rule-based reader & untrained $Y$, FS & 3.0 & 45.1 & 3.3 & 40.7 \\
     & run 1 & $+$11.0 {\scriptsize[$+$3.4, $+$18.6]} & $+$0.4 & $+$13.6 {\scriptsize[$+$7.0, $+$20.1]} & $-$4.0 \\
     & run 2 & $+$9.3 {\scriptsize[$+$2.5, $+$16.0]} & $+$3.0 & $+$11.0 {\scriptsize[$+$4.5, $+$17.4]} & $-$2.6 \\
     & run 3 & $+$4.6 {\scriptsize[$-$1.7, $+$11.0]} & $+$1.7 & $+$13.2 {\scriptsize[$+$6.1, $+$20.7]} & $-$1.5 \\
     & mean & $+$8.3 {\scriptsize[$+$2.0, $+$14.1]} & $+$1.7 & $+$12.6 {\scriptsize[$+$6.9, $+$18.2]} & $-$2.7 \\
    \midrule
    Independent judge & untrained $Y$, FS & 2.1 & 36.3 & 2.2 & 35.9 \\
     & run 1 & $+$8.9 {\scriptsize[$+$0.8, $+$16.9]} & $+$10.1 & $+$7.3 {\scriptsize[$-$0.4, $+$15.2]} & $+$9.9 \\
     & run 2 & $+$6.8 {\scriptsize[$-$0.4, $+$14.3]} & $+$2.5 & $+$11.7 {\scriptsize[$+$3.6, $+$19.3]} & $+$1.8 \\
     & run 3 & $+$5.5 {\scriptsize[$-$2.1, $+$12.8]} & $+$0.4 & $+$16.5 {\scriptsize[$+$8.2, $+$25.1]} & $-$1.5 \\
     & mean & $+$7.0 {\scriptsize[$+$0.4, $+$13.3]} & $+$4.4 & $+$11.8 {\scriptsize[$+$5.2, $+$18.6]} & $+$3.4 \\
    \midrule
    Training judge & untrained $Y$, FS & 4.6 & 46.0 & 4.0 & 49.1 \\
     & run 1 & $+$1.3 {\scriptsize[$-$5.0, $+$7.4]} & $+$21.1 & $+$1.8 {\scriptsize[$-$4.1, $+$7.1]} & $+$16.1 \\
     & run 2 & $+$5.9 {\scriptsize[$+$0.4, $+$11.4]} & $+$10.5 & $+$6.6 {\scriptsize[$+$0.4, $+$12.3]} & $+$7.0 \\
     & run 3 & $+$4.2 {\scriptsize[$-$1.3, $+$9.7]} & $+$4.2 & $+$11.4 {\scriptsize[$+$5.8, $+$17.7]} & $-$2.6 \\
     & mean & $+$3.8 {\scriptsize[$-$1.0, $+$8.2]} & $+$12.0 & $+$6.6 {\scriptsize[$+$2.3, $+$11.0]} & $+$6.8 \\
    \bottomrule
  \end{tabular}
  \caption{Registered endpoints of the replication for each training seed, all entries in percent, on the pairs common to all its endpoint reads, with records read in their original form. Each reader's first row gives the untrained writer's pair-level Youden index $Y$ and false-strike rate FS, and the other rows the paired change from untrained to trained records, with 95\% intervals over patient groups for $\Delta Y$. The mean rows are the seed-averaged changes of the main paper's Table~2.}
  \label{tab:supp_perseed}
\end{table}

Averages over 3 runs can hide 1 run that carries a result, so Figure~\ref{fig:supp_perrun} and Table~\ref{tab:supp_perseed} give the registered endpoints for each training run. The rule-based reader's $Y$ rises for every seed on both populations, and its false-strike rate changes by at most $4.0\%$ for any seed. The training judge's rise in false strikes on the validation pairs follows how many findings each writer lists, $+21.1\%$, $+10.5\%$, and $+4.2\%$ for the writers from runs 1, 2, and 3, which list $7.62$, $10.68$, and $11.87$ of the 12 findings per record on all 476 validation records against $10.43$ for the untrained writer. A set of 3 writers cannot establish a dependence, but the order is the one the main paper's omission account predicts, since a judge reading a record in the original form finds no support for a true denial of a finding the record leaves out, and each such rejection is a false strike.

\section{What the writers put in their records}
\label{sec:supp_records}

\begin{figure*}[t]
  \centering
  \includegraphics[width=\textwidth]{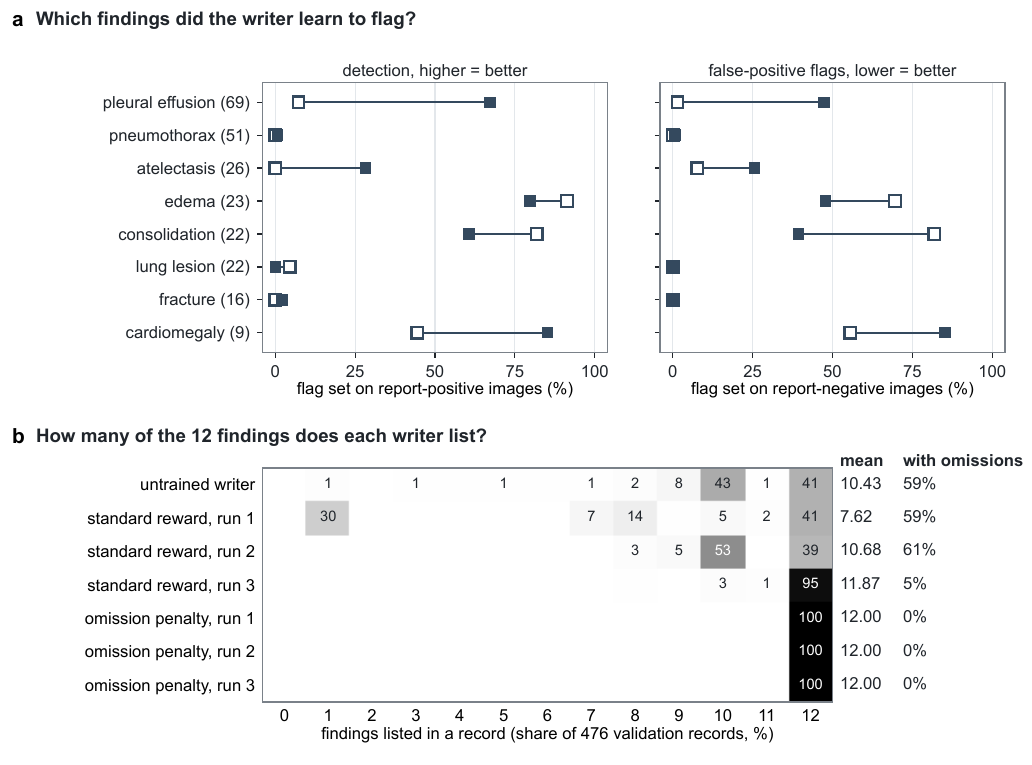}
  \caption{What the writers record, on the validation records, descriptive. (a) For each finding, the share of images whose report label is positive (detection) or negative (false-positive flags) on which the target flag is set, for the untrained writer (open squares) and the mean of the 3 replication runs (filled squares), with the pairs per finding in parentheses, in percent. (b) How many of the 12 findings each writer lists in a record, as the share of its 476 records in each count, with the mean and the share of records that omit at least 1 finding at the right. Listing all 12 findings measures completeness, not correctness.}
  \label{fig:supp_records}
\end{figure*}

The learning is narrow and its form differs by finding (Fig.~\ref{fig:supp_records}a). Pleural effusion is learned by flagging it, with detection rising from $7\%$ to $67\%$ and false-positive flags from $1\%$ to $47\%$, whereas consolidation and edema are learned by flagging less, with false-positive flags falling from $82\%$ to $39\%$ and from $70\%$ to $48\%$. Pneumothorax and fracture stay almost never flagged, lung lesion loses its few flags, and cardiomegaly is flagged on every image in 2 of the 3 runs. How much each writer omits also varies (Fig.~\ref{fig:supp_records}b). The untrained writer omits at least 1 finding on $59\%$ of its records, run 1 of the standard reward lists only 1 finding on $30\%$ of its records, and every record written under the omission penalty lists all 12 findings.

\section{Readers and record ablations}
\label{sec:supp_readers}

The flip rate under a truth-preserving change of wording, which the concurrent study uses to measure wording influence, is not a contamination measure for a record written without the sentence, since a record that is constant across images scores as high on it as the untrained writer, so it serves only as a diagnostic here. The rule-based reader normalizes the record, looks up the entry of the sentence's finding, and supports a sentence that asserts the finding exactly when the flag is present and a sentence that denies it exactly when the flag is absent, and a missing record counts as a rejection. Side, box, note, and wording are ignored. The record ablations change fixed saved records, and each starts from a stated serialization. Original form is the writer's record passed unchanged, as the evaluation procedure does. The reward-scoring form keeps every field and writes every unlisted finding as absent. Sides not applicable starts from the original record, while notes removed and flags only start from the reward-scoring form, so comparing either of them with the original record mixes the ablation with the change of serialization, which is why the main text compares them with the reward-scoring form. Notes removed empties every note and the other field and keeps flags, sides, and boxes. Sides neutral sets every side to not applicable and keeps everything else. Flags only keeps the present flags and sets every side to not applicable, every box to null, and every note and the other field to empty, the same entry format as the records written from the labels. Each condition is applied to both the untrained and the trained records, and the changed records are judged again against the original sentences with the unchanged swap and scoring passes.

\section{Separating omission from order}
\label{sec:supp_omission}

The reward-scoring form changes a record in 2 ways, so 4 further forms separate them, each built from the writer's own entries with the repairs of the reward-scoring form. Order only puts the listed findings in the fixed order and leaves the omitted ones out. Fill only keeps the writer's order and appends every omitted finding as an explicit absent entry. Fill target appends only the omitted target finding, the one the sentence names, and fill other appends every omitted finding except the target. With the original record and the reward-scoring form these give 2 factorial designs, order by fill and target fill by other fill, both read by both judges on the untrained and the trained records. The effect of a factor is its change averaged over the levels of the other factor. The between-judge contrast in the full reward-scoring form was registered before any replicated record was read. The order-by-fill and target-by-other decompositions were both declared after the first replicated run's records had been read in their original and reward-scoring forms, before the intermediate forms were read, the latter with the change in each judge's false-strike increase and the between-judge change in $Y$ when every omission is written as absent in the writer's order as its primary contrasts, so both decompositions are post hoc. Averaged over the 3 runs on the 234 pairs where all its forms parse, stating every omission as absent lowers the training judge's rise in false strikes by $5.7\%$ $[+3.0\%, +8.6\%]$ and the independent judge's by $3.3\%$ $[-0.1\%, +6.5\%]$. A last form, added after the registered results were known and therefore exploratory, appends every omitted finding in the writer's order as an explicit entry whose presence and note read not assessed and whose side is not applicable, so that a judge sees which findings are missing without being told they are absent. In this form, on 236 validation pairs, the training judge's $\Delta Y$ is $+3.7\%$ $[-0.7\%, +8.4\%]$ with a false-strike change of $+12.9\%$ $[+7.9\%, +17.8\%]$, and the independent judge's $+6.6\%$ $[-0.1\%, +13.3\%]$ with $+1.8\%$ $[-3.6\%, +7.8\%]$. Intervals come from 1 patient resampling shared by all quantities of an analysis.

How often those repairs act, and whether they matter, was checked post hoc by replaying the repair step on every record of the registered reads, 1 record per image, with no mismatch against the step itself. Content repairs, such as a dropped duplicate or malformed entry or a dropped box, occur in 6 of 463 untrained validation records, 2 of 463 from run 1, and none from runs 2 and 3, and in at most 4 of 1,016 records per writer on the held-out patients, while side resets, truncated notes, and text flags never occur. Without the 7 validation pairs that contain a repaired record, the difference between the judges stays at $6.2\%$ $[2.1\%, 10.9\%]$ on 230 pairs and the effect of the imputed absences at $4.4\%$ $[0.5\%, 8.4\%]$ on 229 pairs, against $6.2\%$ $[2.0\%, 10.5\%]$ and $4.6\%$ $[0.7\%, 8.6\%]$ on all pairs, so the order and fill factors are not artifacts of the repairs.

\begin{table}[h]
  \centering\footnotesize
  \setlength{\tabcolsep}{3pt}
  \begin{tabular}{llccccc}
    \toprule
    Sentence (pairs) & Judge & Original form & Not assessed & Absent & Marking & Assertion \\
    \midrule
    Denies the finding (148) & Training & $+$15.8 & $+$20.0 & $+$10.6 & $-$4.3 {\scriptsize[$-$6.8, $-$2.0]} & $+$9.5 {\scriptsize[$+$6.0, $+$12.7]} \\
    & Independent & $+$13.3 & $+$9.7 & $+$8.3 & $+$3.6 {\scriptsize[$+$0.5, $+$6.6]} & $+$1.4 {\scriptsize[$-$1.3, $+$4.0]} \\
    Asserts the finding (88) & Training & $+$5.7 & $+$0.8 & $-$0.8 & $+$4.9 {\scriptsize[$+$0.8, $+$9.5]} & $+$1.5 {\scriptsize[$-$1.1, $+$4.9]} \\
    & Independent & $-$11.0 & $-$11.4 & $-$11.7 & $+$0.4 {\scriptsize[$-$5.7, $+$5.4]} & $+$0.4 {\scriptsize[$-$4.1, $+$4.5]} \\
    \bottomrule
  \end{tabular}
  \caption{Exploratory split of the not-assessed contrast by sentence direction, all entries in percent, on the 236 validation pairs common to the original, not-assessed, and absent forms in the writer's order. Entries in the first 3 columns are the seed-averaged change in the false-strike rate from untrained to trained records with the omitted findings left out (in the original form), marked not assessed, or stated absent, all in the writer's order. Marking is original minus not assessed, and assertion is not assessed minus absent, with 95\% intervals over patient groups, where a positive value is a reduction of the false-strike increase.}
  \label{tab:supp_direction}
\end{table}

Table~\ref{tab:supp_direction} shows where the assertion acts. For the training judge it lowers the false-strike increase on denying sentences by $9.5\%$ and gives no detected additional reduction on asserting sentences, and the not-assessed marker alone raises the increase on denials, while on asserting sentences the not-assessed marker lowers it and the further effect of asserting absence is unresolved. The independent judge responds to the marker on denials, and its additional response to the assertion is unresolved, so the judges' responses to the assertion differ by $8.1\%$ $[4.3\%, 11.9\%]$ on denials.

\begin{table}[h]
  \centering\footnotesize
  \setlength{\tabcolsep}{3pt}
  \begin{tabular}{lcccc}
    \toprule
    & \multicolumn{2}{c}{Training judge} & \multicolumn{2}{c}{Independent judge} \\
    Form of untrained and trained held-out test records & $\Delta Y$ & $\Delta$FS & $\Delta Y$ & $\Delta$FS \\
    \midrule
    Original form & $+$6.4 {\scriptsize[$+$2.1, $+$10.8]} & $+$7.2 {\scriptsize[$+$2.2, $+$11.9]} & $+$12.5 {\scriptsize[$+$6.0, $+$18.7]} & $+$3.8 {\scriptsize[$-$2.3, $+$9.5]} \\
    Fixed order, omissions left out & $+$6.7 & $+$6.7 & $+$11.0 & $+$4.7 \\
    Writer's order, omissions absent & $+$11.2 & $+$1.9 & $+$10.5 & $-$0.5 \\
    Reward-scoring form & $+$11.2 {\scriptsize[$+$6.8, $+$16.1]} & $+$2.0 {\scriptsize[$-$3.2, $+$6.9]} & $+$9.6 {\scriptsize[$+$3.7, $+$15.6]} & $-$1.0 {\scriptsize[$-$7.0, $+$4.4]} \\
    \bottomrule
  \end{tabular}
  \caption{Post hoc re-evaluation of the saved held-out test records of the untrained writer and the 3 replication runs, all entries in percent, in the forms of the validation decomposition, with seed-averaged paired changes in the pair-level Youden index ($\Delta Y$) and the false-strike rate ($\Delta$FS), declared before any of these forms was read, on the 270 held-out test pairs where every form parses. The difference between the judges' responses to the whole change of form is $7.7\%$ $[2.8\%, 12.8\%]$, its fill component $6.4\%$ $[2.2\%, 10.6\%]$, and its order component an unresolved $1.3\%$ $[-1.1\%, 3.8\%]$. Stating the omissions as absent lowers both judges' false-strike increase by $5.0\%$ $[2.6\%, 7.3\%]$. The registered held-out test endpoints, on records in the original form, are those of the main paper's Table 2.}
  \label{tab:supp_locked_forms}
\end{table}

\section{The record form under 5 judges}
\label{sec:supp_judges}

\begin{table}[h]
  \centering\scriptsize
  \setlength{\tabcolsep}{3pt}
  \begin{tabular}{lcccccc}
    \toprule
    Judge & Pairs & $\Delta Y$ original & $\Delta Y$ reward-scoring & Form effect, $\Delta Y$ & Fill effect, $\Delta$FS & Minus training judge \\
    \midrule
    \multicolumn{7}{l}{\emph{Validation pairs}} \\
    Qwen3-VL-8B (training) & 236 & $+$3.8 & $+$6.1 & $-$2.3 {\scriptsize[$-$5.2, $+$0.6]} & $+$5.4 {\scriptsize[$+$2.6, $+$8.0]}  & \\
    MedGemma-4B (independent) & 236 & $+$7.2 & $+$3.2 & $+$4.0 {\scriptsize[$-$0.1, $+$8.0]} & $+$3.3 {\scriptsize[$+$0.6, $+$6.2]} & $+$6.2 {\scriptsize[$+$2.0, $+$10.6]} \\
    Qwen3-VL-30B-A3B & 238 & $+$3.9 & $+$6.3 & $-$2.4 {\scriptsize[$-$5.6, $+$1.2]} & $+$6.3 {\scriptsize[$+$3.3, $+$9.0]} & $-$0.1 {\scriptsize[$-$2.7, $+$2.4]} \\
    Phi-4-mini & 238 & $+$0.6 & $+$1.0 & $-$0.4 {\scriptsize[$-$4.8, $+$4.0]} & $+$2.2 {\scriptsize[$-$0.3, $+$4.5]} & $+$1.8 {\scriptsize[$-$2.4, $+$6.1]} \\
    OLMo-2-7B & 238 & $+$4.2 & $+$5.7 & $-$1.5 {\scriptsize[$-$5.8, $+$2.8]} & $+$5.7 {\scriptsize[$+$2.5, $+$8.6]} & $+$0.7 {\scriptsize[$-$3.5, $+$4.7]} \\
    \midrule
    \multicolumn{7}{l}{\emph{Held-out test pairs}} \\
    Qwen3-VL-8B (training) & 270 & $+$6.4 & $+$11.2 & $-$4.8 {\scriptsize[$-$7.8, $-$1.7]} & $+$5.0 {\scriptsize[$+$2.6, $+$7.3]}  & \\
    MedGemma-4B (independent) & 270 & $+$12.5 & $+$9.6 & $+$2.8 {\scriptsize[$-$1.2, $+$7.0]} & $+$5.0 {\scriptsize[$+$2.6, $+$7.3]} & $+$7.7 {\scriptsize[$+$2.8, $+$12.8]} \\
    Qwen3-VL-30B-A3B & 274 & $+$9.4 & $+$13.4 & $-$4.0 {\scriptsize[$-$7.1, $-$1.1]} & $+$6.2 {\scriptsize[$+$3.8, $+$8.4]} & $+$0.7 {\scriptsize[$-$2.1, $+$3.5]} \\
    Phi-4-mini & 274 & $+$5.7 & $+$5.6 & $+$0.1 {\scriptsize[$-$3.0, $+$3.4]} & $+$3.0 {\scriptsize[$+$1.0, $+$5.0]} & $+$4.9 {\scriptsize[$+$0.6, $+$9.6]} \\
    OLMo-2-7B & 274 & $+$9.5 & $+$10.6 & $-$1.1 {\scriptsize[$-$5.7, $+$3.2]} & $+$2.9 {\scriptsize[$-$0.4, $+$6.2]} & $+$3.6 {\scriptsize[$-$0.9, $+$7.7]} \\
    \bottomrule
  \end{tabular}
  \caption{The record-form effect under 5 judges reading the same saved records of the untrained writer and the 3 replication runs, all metric values in percent. $\Delta Y$ is the run-averaged paired change in the pair-level Youden index from untrained to trained records, and the form effect is its value in the original form minus its value in the reward-scoring form, with 95\% intervals over patient groups, so a negative form effect on $\Delta Y$ means the judge registers more of the learning in the reward-scoring form. The fill effect on $\Delta$FS is the seed-averaged change in the false-strike rate with omitted findings left out minus with them written absent, averaged over the 2 orders, so a positive value means that writing the absences lowers the rise in false strikes. The last column is each judge's form effect on $\Delta Y$ minus the training judge's, recomputed on that row's pairs and the same draws, so it need not equal the difference from the training row printed above it. The judges are Qwen3-VL-8B-Instruct, MedGemma-4B-it, Qwen3-VL-30B-A3B-Instruct, Phi-4-mini-instruct, and OLMo-2-1124-7B-Instruct. Pairs are those on which every read of that comparison parses. The 3 further judges were added post hoc, declared before any of their reads, and all rows use 1 analysis script.}
  \label{tab:supp_judges}
\end{table}

The form effect of the main paper compares 2 judges, so 3 further judges, declared post hoc before any of their reads, re-evaluated the same saved validation and held-out test records of the untrained writer and the 3 replication runs in the 4 forms of the order-by-fill design, with the unchanged judge messages and scoring. Qwen3-VL-30B-A3B-Instruct is a mixture-of-experts judge from the training judge's family with about 4 times its total parameters, and Phi-4-mini-instruct and OLMo-2-1124-7B-Instruct are text-only judges from 2 further families. Every verdict of the 3 parsed, and all rows of Table~\ref{tab:supp_judges} come from 1 analysis script with the patient resampling of the main paper.

We observe 2 regularities. Writing omissions as absent, averaged over the 2 orders, lowers the rise in false strikes under all 5 judges in every point estimate, by $2.2\%$ to $6.3\%$ on the validation pairs and $2.9\%$ to $6.2\%$ on the held-out test pairs, resolved in 8 of the 10 cells, all but Phi-4-mini on the validation pairs and OLMo-2 on the held-out test pairs, while the fixed order changes no judge's false strikes by more than $1.4\%$. How much of the learning a judge credits in the reward-scoring form varies across the tested judges. The larger judge of the training judge's family shows no detected difference from the training judge, its form effect differing by $-0.1\%$ $[-2.7\%, +2.4\%]$ and $+0.7\%$ $[-2.1\%, +3.5\%]$, and on the held-out test pairs it too registers a resolved additional gain in the reward-scoring form, $4.0\%$ $[1.1\%, 7.1\%]$. In the reward-scoring form the judges of the other families register a smaller additional gain (OLMo-2), a near-zero and unresolved change (Phi-4-mini), or a smaller gain than in the original form (MedGemma), and their form effects differ from the training judge's in both populations for MedGemma, on the held-out test pairs for Phi-4-mini, by $4.9\%$ $[0.6\%, 9.6\%]$, and by no resolved amount for OLMo-2. Phi-4-mini registers almost none of the learning on the validation pairs in either form. The opposite movement of the 2 judges in the main paper is therefore a contrast between the reward judge and the tested judge that reads the forms most differently. Adding a larger judge of the reward judge's family did not diversify how the forms are read, although 5 judges, 2 from 1 family, cannot separate family from size or training, and the part of the effect shared by every judge tested is the one on false strikes.

\section{The record form under an explicit convention}
\label{sec:supp_conv}

\begin{figure*}[t]
  \centering
  \includegraphics[width=\textwidth]{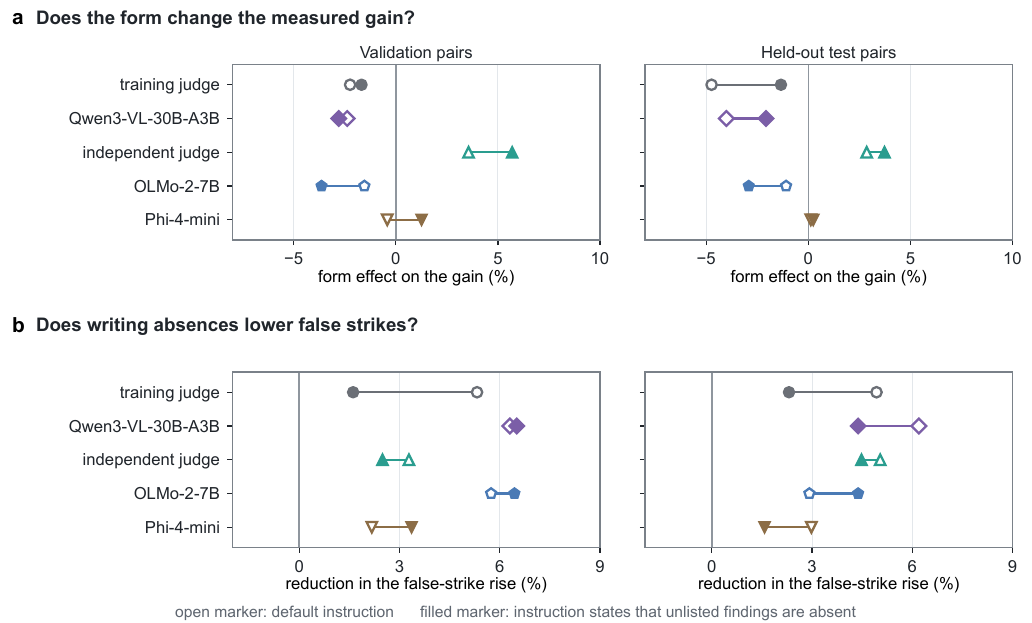}
  \caption{The 5 judges reading the same saved records with their default instruction (open markers) and with 1 added sentence stating that any unlisted finding is absent (filled markers), in percent. (a) Form effect on the gain, the change in $Y$ in the original form minus that in the reward-scoring form. (b) How much writing the omitted findings as absent lowers the rise in false strikes, averaged over the 2 orders. Intervals and pair counts are in Table~\ref{tab:supp_conv}.}
  \label{fig:supp_conv}
\end{figure*}

\begin{table}[h]
  \centering\footnotesize
  \setlength{\tabcolsep}{3pt}
  \begin{tabular}{lccccccc}
    \toprule
    & & \multicolumn{3}{c}{Form effect on $\Delta Y$} & \multicolumn{3}{c}{Fill effect on $\Delta$FS} \\
    Judge & Pairs & Default & Convention & Change & Default & Convention & Change \\
    \midrule
    \multicolumn{8}{l}{\emph{Validation pairs}} \\
    Qwen3-VL-8B (training) & 238 & $-$2.2 & $-$1.7 & $+$0.6 {\scriptsize[$-$1.6, $+$2.5]} & $+$5.3 & $+$1.6 & $-$3.7 {\scriptsize[$-$5.3, $-$2.0]} \\
    Qwen3-VL-30B-A3B & 238 & $-$2.4 & $-$2.8 & $-$0.4 {\scriptsize[$-$2.6, $+$1.7]} & $+$6.3 & $+$6.5 & $+$0.2 {\scriptsize[$-$1.2, $+$2.0]} \\
    MedGemma-4B (independent) & 234 & $+$3.6 & $+$5.7 & $+$2.1 {\scriptsize[$-$0.8, $+$5.5]} & $+$3.3 & $+$2.5 & $-$0.8 {\scriptsize[$-$2.3, $+$0.6]} \\
    OLMo-2-7B & 238 & $-$1.5 & $-$3.6 & $-$2.1 {\scriptsize[$-$5.4, $+$1.6]} & $+$5.7 & $+$6.4 & $+$0.7 {\scriptsize[$-$1.0, $+$2.3]} \\
    Phi-4-mini & 238 & $-$0.4 & $+$1.3 & $+$1.7 {\scriptsize[$-$2.8, $+$6.0]} & $+$2.2 & $+$3.4 & $+$1.2 {\scriptsize[$-$1.4, $+$3.9]} \\
    \multicolumn{8}{l}{\quad MedGemma minus training judge, form effect under the convention, 235 pairs, $+$7.4 {\scriptsize[$+$3.1, $+$12.1]}} \\
    \midrule
    \multicolumn{8}{l}{\emph{Held-out test pairs}} \\
    Qwen3-VL-8B (training) & 274 & $-$4.7 & $-$1.3 & $+$3.4 {\scriptsize[$+$1.1, $+$5.8]} & $+$4.9 & $+$2.3 & $-$2.6 {\scriptsize[$-$4.2, $-$1.1]} \\
    Qwen3-VL-30B-A3B & 274 & $-$4.0 & $-$2.1 & $+$1.9 {\scriptsize[$-$0.4, $+$4.5]} & $+$6.2 & $+$4.4 & $-$1.8 {\scriptsize[$-$3.4, $-$0.3]} \\
    MedGemma-4B (independent) & 268 & $+$2.9 & $+$3.7 & $+$0.9 {\scriptsize[$-$2.6, $+$4.3]} & $+$5.0 & $+$4.5 & $-$0.6 {\scriptsize[$-$2.1, $+$1.1]} \\
    OLMo-2-7B & 274 & $-$1.1 & $-$2.9 & $-$1.8 {\scriptsize[$-$5.3, $+$1.9]} & $+$2.9 & $+$4.4 & $+$1.5 {\scriptsize[$-$1.0, $+$3.9]} \\
    Phi-4-mini & 274 & $+$0.1 & $+$0.2 & $+$0.1 {\scriptsize[$-$3.5, $+$3.5]} & $+$3.0 & $+$1.6 & $-$1.4 {\scriptsize[$-$3.8, $+$0.9]} \\
    \multicolumn{8}{l}{\quad MedGemma minus training judge, form effect under the convention, 272 pairs, $+$5.0 {\scriptsize[$-$0.4, $+$10.7]}} \\
    \bottomrule
  \end{tabular}
  \caption{The same saved records read by each judge with its default instruction and with 1 added sentence stating that any of the 12 findings the record does not list is absent, all metric values in percent, a post hoc control declared before any of its reads. Form and fill effects are defined as in Table~\ref{tab:supp_judges}, recomputed here on the pairs where every read of the judge under both instructions parses (the between-judge rows use the pairs where both judges parse under the convention), and the change is the value under the convention minus the value under the default instruction, with 95\% intervals over patient groups.}
  \label{tab:supp_conv}
\end{table}

If a judge treated a stated convention as it treats written entries, telling it what an omission means would make the original record and the reward-scoring form carry the same meaning, and writing the absences explicitly would have no additional effect. A post hoc control, declared before any of its reads, tests this by re-evaluating the same saved records with all 5 judges in the 4 forms, with 1 sentence appended to every judge message in the record and swap passes, stating that any of the 12 findings the record does not list is absent. The rest of every message and the scoring are unchanged, and a check run confirmed that the sentence reached every call.

The reward judge follows the convention in part (Table~\ref{tab:supp_conv}). Its reliance on written absences for false strikes falls from $5.3\%$ to $1.6\%$ on the validation pairs and from $4.9\%$ to $2.3\%$ on the held-out test pairs, both changes resolved, and on the held-out test pairs most of its form effect on the gain disappears, from $-4.7\%$ to $-1.3\%$. The larger judge of its family responds on the held-out test pairs only, with a resolved change for false strikes. MedGemma, OLMo-2, and Phi-4-mini show no resolved response in either population, so for them no change under the convention is detected. With both told the convention, MedGemma and the training judge still differ by $7.4\%$ $[3.1\%, 12.1\%]$ on the validation pairs, and by an unresolved $5.0\%$ $[-0.4\%, 10.7\%]$ on the held-out test pairs. A stated convention is therefore a partial repair that works for the judges that follow it, here the reward judge's family, and writing explicit absences lowers the false-strike increase in point estimate for every tested judge under both instructions.

\section{Judge size, judge instruction, and silence}
\label{sec:supp_judgevar}

\begin{table*}[t]
  \centering\scriptsize
  \setlength{\tabcolsep}{3pt}
  \begin{tabular}{llcccc}
    \toprule
    Comparison & Population & Pairs & Form effect, judge B & Form effect, judge A & B minus A \\
    \midrule
    Gemma-3-4B vs training judge & validation & 238 & $-$1.7 {\scriptsize[$-$5.8, $+$2.5]} & $-$2.2 {\scriptsize[$-$5.0, $+$0.7]} & $+$0.6 {\scriptsize[$-$4.0, $+$4.8]} \\
    Gemma-3-4B vs training judge & held-out test & 274 & $+$0.1 {\scriptsize[$-$4.9, $+$5.1]} & $-$4.7 {\scriptsize[$-$8.1, $-$1.7]} & $+$4.9 {\scriptsize[$-$0.2, $+$10.2]} \\
    Gemma-3-12B vs training judge & validation & 238 & $-$0.7 {\scriptsize[$-$4.1, $+$2.8]} & $-$2.2 {\scriptsize[$-$5.0, $+$0.7]} & $+$1.5 {\scriptsize[$-$1.6, $+$4.7]} \\
    Gemma-3-12B vs training judge & held-out test & 274 & $-$1.6 {\scriptsize[$-$4.0, $+$0.6]} & $-$4.7 {\scriptsize[$-$8.1, $-$1.7]} & $+$3.2 {\scriptsize[$+$0.1, $+$6.3]} \\
    Gemma-3-27B vs training judge & validation & 238 & $-$2.0 {\scriptsize[$-$5.2, $+$1.3]} & $-$2.2 {\scriptsize[$-$5.0, $+$0.7]} & $+$0.3 {\scriptsize[$-$2.4, $+$3.0]} \\
    Gemma-3-27B vs training judge & held-out test & 274 & $-$2.8 {\scriptsize[$-$6.1, $+$0.5]} & $-$4.7 {\scriptsize[$-$8.1, $-$1.7]} & $+$1.9 {\scriptsize[$-$1.1, $+$5.0]} \\
    \midrule
    Training judge, cite vs default & validation & 238 & $-$3.9 {\scriptsize[$-$7.7, 0.0]} & $-$2.2 {\scriptsize[$-$5.0, $+$0.7]} & $-$1.7 {\scriptsize[$-$4.4, $+$1.1]} \\
    Training judge, cite vs default & held-out test & 274 & $-$4.6 {\scriptsize[$-$8.2, $-$1.2]} & $-$4.7 {\scriptsize[$-$8.1, $-$1.7]} & $+$0.1 {\scriptsize[$-$2.9, $+$3.1]} \\
    Training judge, reason first vs default & validation & 235 & $-$4.1 {\scriptsize[$-$7.3, $-$1.1]} & $-$2.4 {\scriptsize[$-$4.9, $+$0.1]} & $-$1.7 {\scriptsize[$-$4.8, $+$1.3]} \\
    Training judge, reason first vs default & held-out test & 274 & $-$1.8 {\scriptsize[$-$5.2, $+$1.6]} & $-$4.7 {\scriptsize[$-$8.1, $-$1.7]} & $+$2.9 {\scriptsize[$-$0.2, $+$6.3]} \\
    MedGemma-4B, cite vs default & validation & 232 & $-$1.3 {\scriptsize[$-$5.7, $+$3.2]} & $+$4.0 {\scriptsize[$-$0.1, $+$8.3]} & $-$5.3 {\scriptsize[$-$10.7, 0.0]} \\
    MedGemma-4B, cite vs default & held-out test & 266 & $-$0.9 {\scriptsize[$-$4.7, $+$3.0]} & $+$2.6 {\scriptsize[$-$1.5, $+$6.7]} & $-$3.5 {\scriptsize[$-$9.0, $+$2.0]} \\
    MedGemma-4B, reason first vs default & validation & 222 & $-$3.5 {\scriptsize[$-$8.3, $+$1.1]} & $+$3.5 {\scriptsize[$-$0.9, $+$7.7]} & $-$6.9 {\scriptsize[$-$12.2, $-$1.5]} \\
    MedGemma-4B, reason first vs default & held-out test & 250 & $-$3.1 {\scriptsize[$-$7.2, $+$1.4]} & $+$3.7 {\scriptsize[$-$0.7, $+$8.4]} & $-$6.8 {\scriptsize[$-$12.5, $-$1.2]} \\
    \midrule
    MedGemma-4B vs training judge, both cite & validation & 234 & $-$1.3 {\scriptsize[$-$5.6, $+$3.4]} & $-$4.1 {\scriptsize[$-$8.1, $-$0.1]} & $+$2.8 {\scriptsize[$-$2.9, $+$8.5]} \\
    MedGemma-4B vs training judge, both cite & held-out test & 270 & $-$1.2 {\scriptsize[$-$5.0, $+$2.6]} & $-$4.4 {\scriptsize[$-$8.2, $-$0.9]} & $+$3.2 {\scriptsize[$-$2.2, $+$9.4]} \\
    MedGemma-4B vs training judge, both reason first & validation & 221 & $-$3.0 {\scriptsize[$-$7.1, $+$1.1]} & $-$4.8 {\scriptsize[$-$7.8, $-$1.9]} & $+$1.8 {\scriptsize[$-$2.7, $+$6.3]} \\
    MedGemma-4B vs training judge, both reason first & held-out test & 254 & $-$3.1 {\scriptsize[$-$7.0, $+$1.1]} & $-$1.7 {\scriptsize[$-$5.5, $+$2.2]} & $-$1.4 {\scriptsize[$-$6.9, $+$4.1]} \\
    \bottomrule
  \end{tabular}
  \caption{Judge size and judge instruction, in percent, all reads declared post hoc before they were made. The form effect is $\Delta Y$ in the original form minus $\Delta Y$ in the reward-scoring form, each $\Delta Y$ comparing trained with untrained records, as in Table~\ref{tab:supp_judges}, with 95\% intervals over patient groups. In the instruction rows judge A reads with its default instruction and judge B with the named one. Pair counts differ because a pair enters only when every read of both judges returns a supported or unsupported verdict on both of its images.}
  \label{tab:supp_judgevar}
\end{table*}

\begin{table*}[t]
  \centering\scriptsize
  \setlength{\tabcolsep}{2.5pt}
  \begin{tabular}{llcccccc}
    \toprule
    & & \multicolumn{4}{c}{Target finding not mentioned by the record} & \multicolumn{2}{c}{Target finding listed} \\
    Judge & Population & Correct negative & Incorrect negative & Correct positive & Incorrect positive & Correct negative & Incorrect negative \\
    \midrule
    Training judge & validation & 64.0$\to$99.0 & 65.8$\to$97.5 & 6.6$\to$6.6 & 4.8$\to$8.3 & 64.1$\to$64.1 & 52.4$\to$52.6 \\
    MedGemma-4B & validation & 79.0$\to$99.0 & 71.8$\to$97.4 & 15.8$\to$31.6 & 24.1$\to$39.8 & 67.9$\to$72.6 & 56.3$\to$61.3 \\
    Qwen3-VL-30B & validation & 40.0$\to$99.0 & 44.3$\to$96.2 & 6.6$\to$6.6 & 4.8$\to$8.3 & 63.5$\to$63.7 & 52.4$\to$52.6 \\
    Phi-4-mini & validation & 97.0$\to$99.0 & 83.5$\to$97.5 & 17.1$\to$14.5 & 17.9$\to$16.7 & 86.7$\to$91.3 & 81.6$\to$88.2 \\
    OLMo-2-7B & validation & 45.0$\to$97.0 & 49.4$\to$89.9 & 3.9$\to$18.4 & 3.6$\to$14.3 & 64.9$\to$65.1 & 53.6$\to$54.4 \\
    Gemma-3-4B & validation & 1.0$\to$80.0 & 1.3$\to$74.7 & 5.3$\to$2.6 & 2.4$\to$1.2 & 52.4$\to$52.8 & 45.6$\to$45.8 \\
    Gemma-3-12B & validation & 91.0$\to$100.0 & 79.7$\to$97.5 & 18.4$\to$9.2 & 17.9$\to$9.5 & 64.1$\to$63.9 & 52.4$\to$52.4 \\
    Gemma-3-27B & validation & 56.0$\to$99.0 & 54.4$\to$96.2 & 10.5$\to$9.2 & 9.5$\to$10.7 & 63.5$\to$63.5 & 52.2$\to$52.6 \\
    \midrule
    Training judge & held-out test & 53.5$\to$96.1 & 60.8$\to$91.5 & 5.6$\to$11.1 & 3.2$\to$3.2 & 60.3$\to$60.6 & 46.3$\to$46.8 \\
    MedGemma-4B & held-out test & 74.0$\to$96.9 & 75.4$\to$94.6 & 14.8$\to$33.3 & 17.7$\to$33.9 & 66.8$\to$71.0 & 53.4$\to$58.0 \\
    Qwen3-VL-30B & held-out test & 46.5$\to$95.3 & 54.6$\to$90.8 & 5.6$\to$14.8 & 4.8$\to$8.1 & 61.0$\to$61.1 & 46.6$\to$46.6 \\
    Phi-4-mini & held-out test & 87.4$\to$96.9 & 82.3$\to$94.6 & 14.8$\to$18.5 & 17.7$\to$8.1 & 85.6$\to$90.2 & 78.3$\to$83.6 \\
    OLMo-2-7B & held-out test & 54.3$\to$94.5 & 64.6$\to$90.0 & 11.1$\to$14.8 & 4.8$\to$11.3 & 62.2$\to$64.6 & 49.8$\to$51.6 \\
    Gemma-3-4B & held-out test & 1.6$\to$63.8 & 3.1$\to$63.8 & 5.6$\to$5.6 & 4.8$\to$4.8 & 43.4$\to$45.1 & 33.4$\to$35.4 \\
    Gemma-3-12B & held-out test & 91.3$\to$95.3 & 80.0$\to$91.5 & 16.7$\to$16.7 & 12.9$\to$9.7 & 62.4$\to$62.2 & 48.9$\to$49.2 \\
    Gemma-3-27B & held-out test & 50.4$\to$94.5 & 57.7$\to$90.0 & 5.6$\to$16.7 & 4.8$\to$8.1 & 62.4$\to$62.9 & 48.2$\to$48.4 \\
    \bottomrule
  \end{tabular}
  \caption{What each judge does with silence, default instruction, in percent: the share of verdicts accepting the report sentence on the record as written $\to$ with every omission written absent, for the same saved records of the untrained writer and the 3 replication runs. Negative and positive denote the target finding's absence and presence as assigned by the pair's report-derived labels, correct and incorrect denote agreement with those labels, and the target finding is the sentence's. Each verdict is weighted equally, pooled over the 4 record sets, and a parsed verdict that is neither supported nor unsupported counts as not accepted. Verdicts per cell, in column order, for the training judge (other judges at most 1 fewer through unparsed replies): validation 100, 79, 76, 84, 496, 517; held-out test 127, 130, 54, 62, 625, 622. Differences with 95\% intervals over patient groups are in the released result files.}
  \label{tab:supp_silence}
\end{table*}

The between-judge contrast could reflect the size of the independent judge, its medical tuning, or the instruction both judges read, so the same saved records of the untrained writer and the 3 replication runs were reread, in the 4 forms and in both populations, by 3 general Gemma-3 judges \cite{gemma3_2025} of 4, 12, and 27 billion parameters (the last on 2 GPUs) with the unchanged judge messages, and by both primary judges under 2 instructions that change only the requested reply, 1 asking for the record entry about the sentence's finding, quoted, or the words not listed, and 1 asking for 2 or 3 sentences of reasoning before the verdict. Every read was declared post hoc before it was made, models were loaded at fixed revisions, and the analysis code is the one used for Table~\ref{tab:supp_judges}. Every reply parsed except 99 of 15{,}232 and 117 of 17{,}536 of MedGemma's under the reasoning instruction, and 10 parsed verdicts of the training judge under that instruction and 4 of MedGemma's under the instruction to cite were neither supported nor unsupported; a pair enters a contrast only when every read of both judges gives 1 of the 2 verdicts on both of its images. The medical model at 27 billion parameters was not available to us, so size and medical tuning are compared with 1 model per point and cannot be separated (Table~\ref{tab:supp_judgevar}).

Table~\ref{tab:supp_silence} shows, for the default instruction, what each judge does when the record says nothing about the sentence's finding, on the report sentences of the same saved records. Judges range from accepting almost no label-consistent negative statement about such a finding to accepting almost all, writing every omission as absent moves all of them toward acceptance of negative statements, label-consistent or not, and MedGemma and OLMo-2, with Gemma-3-27B on the correct positive statements of the held-out test records, also accept more positive statements that the record, once filled, contradicts. These rates are descriptive. They pool the matched report-sentence verdicts on the records of the untrained writer and the 3 runs, each verdict weighted equally, so record sets with more omitted targets weigh more, and patient-group resampling keeps the dependence across record sets and forms. Writing every omission absent also changes the record beyond the target, so the transitions do not isolate the target's silence.

Under the instruction to cite, on records in their original form, the training judge reported the target finding as not listed in 570 of 730 verdicts whose record omits it ($78.1\%$) on the validation records and 578 of 774 ($74.7\%$) on the held-out test records, pooling the 4 record sets and the report and swapped sentences, whereas MedGemma reported none as not listed and returned the sentence under test itself as its evidence in 306 of 730 ($41.9\%$) and 310 of 774 ($40.1\%$) of those verdicts, and in 928 of 3{,}078 ($30.1\%$) and 792 of 3{,}610 ($21.9\%$) of the verdicts whose record lists the finding. The evidence a judge writes under an instruction to cite need not be the evidence it uses, and this instruction itself changes MedGemma's form effect, so these rates describe the instructed behaviour only. The rereads of the held-out test records in this section drew the sentences of their swapped claims from 200 pairs per finding and direction rather than the 40 of the earlier held-out reads, which changes only those swapped sentences. Every $Y$, false-strike rate, form effect, and contrast, and a pair's eligibility for them, is computed from the report sentences alone, so Table~4 of the paper, Table~\ref{tab:supp_judgevar}, and Table~\ref{tab:supp_silence} are unaffected, and the swapped sentences of the revised setting enter only the held-out citation counts above.

\section{Second writer}
\label{sec:supp_writer2}

As a generality check registered before it ran, Qwen2.5-VL-7B-Instruct was trained with the unchanged recipe, the same frozen judge, and 1 seed, with the adapter restricted to its language layers, since its vision blocks share the projection names. Its records were written by the trained writer and then judged by the unchanged training judge, by the independent judge, and by the rule-based reader. No discrimination gain was detected. The rule-based reader's gain was $+2.1\%$ $[-3.3\%, +7.7\%]$, and its records, which listed 2.2 findings at the start, listed exactly 1 after training, the finding it marked present, leaving all others out.

\section{Omission-cost follow-up}
\label{sec:supp_omitcost}

A follow-up registered after the first replicated run showed the omissions, and before any of its own training, repeats the replication with 1 change to the reward, which becomes the content reward minus $0.25$ times the fraction of the 12 findings the writer's record leaves out, floored at zero. The judge still reads the reward-scoring form. Its endpoints are the replication's unchanged joint criterion on records in the original form, the paired change relative to the matching replicated run for each reader, and the findings listed and target omissions by sentence direction, with 1 read of the held-out test set after the validation endpoints. All 3 runs reached 40 steps. On the 237 validation pairs common to the omission-cost endpoint reads, a set that differs from that of Table 2 of the paper by 1 pair, the rule-based reader's seed-averaged gain was $+6.9\%$ $[+1.2\%, +12.1\%]$ with a false-strike change of $+5.2\%$ (one-sided upper bound $8.9\%$), the independent judge's $+5.3\%$ $[-1.2\%, +11.4\%]$ with $+6.8\%$ ($10.8\%$), and the training judge's $+4.6\%$ $[+0.4\%, +8.9\%]$ with $+6.3\%$ ($9.3\%$), so the joint criterion is not met. Paired against the matching replicated runs on 236 pairs, the changes in $Y$ were $-1.4\%$ $[-5.1\%, +2.1\%]$, $-2.0\%$ $[-6.4\%, +2.1\%]$, and $+0.7\%$ $[-2.1\%, +3.6\%]$ for the rule-based reader, the independent judge, and the training judge, and the changes in the false-strike rate $+3.5\%$ $[+0.1\%, +6.9\%]$, $+2.4\%$ $[-1.1\%, +6.0\%]$, and $-5.5\%$ $[-8.4\%, -2.7\%]$. The rule-based reader's rise comes from 1 run, $+12.7\%$ against $-1.3\%$ and $-0.8\%$. Every omission-cost record lists all 12 findings, $1.953$ $[+1.812, +2.085]$ more than its control. In a descriptive count, the run whose false strikes under the rule-based reader rose flags the denied target present on 69 of 149 true images, against 42 for its control, while the other 2 mark fewer findings present than their controls, 4.34 and 3.29 per record against 5.64 and 4.10. On the 274 held-out test pairs, read once after these results, the rule-based reader's gain was $+11.2\%$ $[+6.1\%, +16.7\%]$ with a false-strike change of $+0.6\%$ (one-sided upper bound $4.4\%$), the independent judge's $+10.5\%$ $[+4.1\%, +17.2\%]$ with $+5.5\%$ ($9.9\%$), and the training judge's $+6.9\%$ $[+2.5\%, +11.3\%]$ with $+0.6\%$ ($4.3\%$), so the criterion is again not met through the independent judge's false strikes, as for the replication's held-out test read. Paired against the controls on 273 held-out test pairs, the changes in $Y$ were $-1.3\%$ $[-4.9\%, +2.2\%]$, $-1.5\%$ $[-5.8\%, +2.9\%]$, and $+0.4\%$ $[-2.7\%, +3.4\%]$, and in the false-strike rate $+3.3\%$ $[+0.4\%, +6.3\%]$, $+1.8\%$ $[-1.5\%, +5.3\%]$, and $-6.2\%$ $[-9.2\%, -3.4\%]$, with $1.81$ $[+1.71, +1.92]$ more findings listed.

Charged for omitting, the writers state every finding without a detected improvement in discrimination, 1 of them flagging the denied finding present on 46.3\% of the 149 true images of denying sentences against 28.2\% for its control, so a complete record is not by itself a correct one.

\section{Records and the judges' reasons}
\label{sec:supp_qual}

The records below were chosen by rules declared before any candidate was looked at, from the validation pairs in list order with their original sentences. They illustrate the analyses and add no evidence to them, and every reason is quoted as the judge wrote it.

The record in Figure~1 of the paper comes from the first pleural-effusion pair with a denying sentence whose run-1 record of the report image leaves the finding out and which the training judge rejects in the original form and accepts in the reward-scoring form. The sentence is ``No pleural effusions.'', true for this image, and the original record is the following.

\begin{Verbatim}[breaklines,fontsize=\footnotesize]
{"record": [{"finding": "Enlarged Cardiomediastinum", "present": true, "side": "not applicable", "box": null,
  "note": "Cardiac silhouette appears widened relative to thoracic width."}],
 "other": "No other abnormalities are visible in this radiograph."}
\end{Verbatim}

\begin{table}[h]
  \centering\footnotesize
  \begin{tabular}{llp{0.62\linewidth}}
    \toprule
    Judge & Form & Verdict and reason \\
    \midrule
    Training & in the original form & rejected, ``The record does not mention pleural effusions, so the sentence cannot be confirmed as true based on available data.'' \\
    Training & reward-scoring form & accepted, ``Pleural Effusion is marked as absent in the record.'' \\
    Independent & in the original form & rejected, ``The record does not mention pleural effusions.'' \\
    Independent & reward-scoring form & accepted, ``The record indicates that Pleural Effusion is not present.'' \\
    \bottomrule
  \end{tabular}
  \caption{The 2 judges on the record above in its 2 forms. The rule-based reader, which normalizes every record, accepts the sentence in either form.}
  \label{tab:supp_qual_flip}
\end{table}

For this record both judges' quoted reasons refer to the effusion entry, missing in the original form and absent in the reward-scoring form, and not to the free-text line (Table~\ref{tab:supp_qual_flip}). That line states that no other abnormality is visible, yet in the original form both judges reject the denial, and once the reward-scoring form states the absence both accept. The omission-cost writer with the same seed writes a record of this image that lists all 12 findings, every one absent and every side, note, and the other field set to not applicable, including the enlarged cardiomediastinum its matched control marked present, and the training judge accepts the sentence in the original form because ``The record indicates no pleural effusion is present.''

The rule for an asserting pleural-effusion sentence on which the run-1 records turn an effusion the untrained writer did not flag into a flag that all 3 readers credit on both images found no pair. Only 7 of the 69 validation pleural-effusion pairs carry an asserting sentence, 2 of them meet the conditions on the flags, and on each of these the training judge is wrong on at least 1 image.

The first cardiomegaly pair whose run-2 records flag cardiomegaly on both images has a detailed sentence (Sec.~\ref{sec:supp_split}) reporting mild enlargement of the cardiac silhouette together with a prominent left atrium. Both records flag cardiomegaly present with the same note, ``Heart size exceeds half the thoracic diameter.'', so the rule-based reader accepts the sentence on both images and the pair adds nothing to its $Y$. The training judge rejects it on both images because neither record mentions a prominent left atrium. The independent judge rejects it on the report image for the same reason and accepts it on the other image, where it reasons that the flagged findings ``can be associated with left atrial enlargement''. On this pair the flags cannot separate the images, and what each judge makes of the sentence's detail decides its verdicts.

\section{Sentence split}
\label{sec:supp_split}

The split was fixed before it was computed. A sentence is detailed if its lower-cased text contains any listed word as a whole word, and plain otherwise. The severity and size words are mild, mildly, moderate, moderately, severe, severely, small, smaller, large, larger, trace, minimal, minimally, tiny, subtle, slight, slightly, marked, markedly, extensive, massive, significant, borderline. The location and laterality words are left, right, bilateral, bilaterally, basilar, bibasilar, base, bases, basal, apex, apical, upper, lower, middle, mid, lobe, lingula, lingular, retrocardiac, perihilar, hilar, costophrenic, lateral, medial, anterior, posterior. The change words are increased, increasing, decreased, decreasing, improved, improving, worsened, worsening, unchanged, stable, new, persistent, resolved, resolving, interval. Hedges were counted and not used, namely possible, possibly, likely, may, might, could, questionable, suspicious, probable, suggest, suggests, suggesting. On the 238 validation pairs the rule gives 144 plain and 94 detailed sentences.

\section{Direct supervision}
\label{sec:supp_sft}

The supervised writers use the same base model, adapter, and writer prompt. The target is the record the image's label vector implies, 12 entries in the fixed order with present equal to a positive label, side not applicable, box null, and empty note and other field. Training runs 1 epoch with micro-batch 1 and gradient accumulation 8, the AdamW optimizer (Adam with decoupled weight decay) at learning rate $10^{-4}$ with 3\% linear warm-up and cosine decay, gradient clipping at 1.0, and random seed 101, with the loss on the answer tokens only. The target variant masks every answer token outside the entries of the findings named by the image's training sentences. Under teacher forcing the entries listed before a target entry, with their label-derived flags, still precede its supervised tokens in the context, so this variant withholds the supervision of the other flags but not all of their presence in the input, a difference from the reinforcement run, whose writer only ever conditions on its own records. The matched variants use the 1,331 images on which the reinforcement run computed training rewards. Each variant was trained once with its settings fixed before it ran, and the weighted variant was specified after inspecting the first supervised result on the validation pairs. After the first supervised result, in which the writer marked nothing present, a weighted variant declared before the remaining results restricts the loss to the 12 tokens that carry the present values, weights a positive value by the matched set's ratio of negative to positive flags, 6.35, and trains for 3 epochs on the 1,331 matched images, with all other settings unchanged.

All supervised writers were read on the 238 validation pairs like the reinforcement writers. Trained on the whole label-derived record, the writer marks nothing present on 432 of 476 records after the 1,331 matched images and on 439 after all 4,501 training images, and the rule-based reader's gain is $+0.0\%$ $[-5.2\%, +5.5\%]$ and $+0.4\%$ $[-4.6\%, +5.5\%]$, although the labels carry 1.63 positive findings per image on average and only 17.7\% of images have none. Supervised only on the entries of the sentences' findings, the writer marks all 12 findings present on every record, and the training judge's $Y$ changes by $-4.6\%$ $[-8.4\%, -0.8\%]$. The weighted variant learns, with a gain under the rule-based reader of $+15.1\%$ $[+7.3\%, +22.6\%]$ at a false-strike increase of $7.6\%$ (one-sided upper bound $13.8\%$), gains of $+6.7\%$ $[-2.6\%, +15.8\%]$ and $+7.1\%$ $[-0.4\%, +15.1\%]$ under the independent and training judges, and a target-flag true-positive rate minus false-positive rate of 18\% against 10\% for the development reinforcement run. Each supervised writer was trained once, so these are single-run development comparisons, not a matched replicated contrast.

\section{Photograph study with single-object claims}
\label{sec:supp_photo}

\begin{table*}[t]
  \centering\footnotesize
  \setlength{\tabcolsep}{4pt}
  \begin{tabular}{lccccc}
    \toprule
    & \multicolumn{2}{c}{$Y$, older / newer writer} & Omission form & Encoding & Comprehension \\
    Judge & blank & absent & absent $-$ blank & sparse $-$ complete & complete / sparse \\
    \midrule
    Qwen3-VL-8B (training judge's model) & 90.0 / 88.8 & 91.6 / 89.4 & $-$1.0 {\scriptsize[$-$2.5, $+$0.6]} & $-$0.1 {\scriptsize[$-$1.3, $+$1.1]} & 100 / 100 \\
    Gemma-3-12B & 89.7 / 87.4 & 91.8 / 89.1 & $-$0.3 {\scriptsize[$-$2.2, $+$1.6]} & $+$0.1 {\scriptsize[$-$1.1, $+$1.3]} & 100 / 100 \\
    \midrule
    Qwen3-VL-30B-A3B & 88.3 / 88.0 & 91.9 / 89.7 & $-$1.9 {\scriptsize[$-$3.3, $-$0.4]} & $+$0.3 {\scriptsize[$-$0.7, $+$1.2]} & 100 / 97.9 \\
    Phi-4-mini & 85.8 / 85.5 & 63.8 / 69.3 & $+$5.8 {\scriptsize[$+$3.0, $+$8.5]} & $-$6.8 {\scriptsize[$-$9.7, $-$3.9]} & 87.5 / 100 \\
    OLMo-2-7B & 61.6 / 60.5 & 60.1 / 56.5 & $-$2.5 {\scriptsize[$-$5.7, $+$0.6]} & $-$2.8 {\scriptsize[$-$6.0, $+$0.3]} & 87.5 / 91.7 \\
    \midrule
    Between the primary judges (registered) & & & $-$0.7 {\scriptsize[$-$2.1, $+$0.7]} & $-$0.2 {\scriptsize[$-$1.2, $+$0.8]} & \\
    \bottomrule
  \end{tabular}
  \caption{The photograph study, read once under the frozen protocol, all entries in percent; study outcomes use the 510 writer-complete pairs and comprehension 48 synthetic records per encoding. $Y$ is the pair-level Youden index of each untrained writer's sparse-style records under each judge with omitted objects left blank or written as absent. The omission-form column gives the change in the writer difference, newer minus older, from blank to absent, and the encoding column the change from the complete state list to the present objects alone under the stated convention, with 95\% intervals over 10,000 resamples of pairs. Comprehension is the share of 48 synthetic records per encoding read as the convention implies, with 90\% required; the 2 judges below it are secondary and reported, not replaced. The last row holds the 2 registered estimands, each requiring an interval that excludes zero and an absolute estimate of at least 5\%, computed from unrounded values.}
  \label{tab:supp_photo}
\end{table*}

The study asks whether the between-judge form contrast of the main paper is a property of the judge families or of the radiology records. It was registered as an inference-only study, and its protocol, pool, labels, screen outputs, code, and pinned model revisions were frozen under a byte-level hash manifest after the automated screen and before any writer record or judge read of a study image; the study jobs verify the manifest before any model loads, load every model at its pinned revision, write each output once, and refuse to overwrite it. The analysis ran once.

\paragraph{Pairs and truth.} The pool holds 516 image-disjoint pairs, 1,032 COCO 2014 images \cite{lin2014coco}, from the official complementary pairs of VQA v2 \cite{goyal2017making} whose question asks whether an object is in the image. A pair is kept when at least 8 of 10 annotators give opposite answers on its 2 images and the COCO instance annotations agree with those answers, and each of 20 object categories contributes at most 60 pairs. Truth is the agreement of these 2 annotation sources, not independently verified truth. Before any writer record, a screening model from a third family, Idefics3-8B-Llama3, answered whether the object is in each image, blind to labels and records; it agreed on 97.8\% of images, and 493 pairs pass. The primary population is all 516 pairs, so no model-selected subset carries the claim, and the 493 screened pairs are a secondary population.

\paragraph{Writers, forms, and encodings.} Qwen3-VL-8B-Instruct and Qwen2.5-VL-7B-Instruct, untrained, each write 1 record per image at temperature 0, without the question, the claim, or the partner image, in a sparse style that lists the objects seen from the 20 with a short note each, and in a secondary checklist style that states every object present or not. A reply that stops at the 2,048-token limit is never a record; in the sparse style 3 replies of each writer did, leaving 510 writer-complete pairs, the 6 missing pairs being 1 bird, 1 boat, 2 car, 1 person, and 1 umbrella pair, and in the secondary checklist style 3 replies of Qwen2.5-VL-7B and none of Qwen3-VL-8B did, leaving 513 pairs for that style. One renderer builds every form from the decoded record with a fixed object order and never sees the claim. In the omission-form arm the judge receives its default instruction and the record with omitted objects left out, written as absent, or written as not reported. In the encoding arm every message carries the same convention, ``Convention for this record: the listed states are authoritative for whether an object is in the photograph. Any object that the record does not list is absent from the photograph, even though the record does not say so explicitly. The listed states and this default take precedence over the notes and the other field.'', and the record is written either as all 20 states or as the present objects alone, a deterministic decoder asserting before every judge call that the 2 encodings decode to the same states and notes. Control records set only the focal object's state to the truth or to its negation, with notes removed, and test whether a judge reads the stated fact.

\paragraph{Judges and criteria.} The primary pair is the training judge's model, Qwen3-VL-8B-Instruct, and Gemma-3-12B-it \cite{gemma3_2025}, chosen before any read for continuity with the paper's independent judge; Qwen3-VL-30B-A3B, Phi-4-mini, and OLMo-2-7B form a secondary panel and no judge is replaced after any output. Each judge reads the record and the claim as text at temperature 0 and must answer with 1 JSON object naming a verdict and a reason. Before the study reads, 48 synthetic records per encoding checked that each judge follows the convention; the criterion is 90\%, and Phi-4-mini and OLMo-2 failed it on complete records at 87.5\% each while both primary judges scored 100\%. On the study records the target controls agreed with the record-implied verdict in at least 99.6\% of the verdicts of each primary judge, writer, encoding, and control cell.

\paragraph{Estimands and result.} With $G_{j,f}$ the difference in $Y$ between the newer and the older writer under judge $j$ and form $f$, the omission-form estimand is the difference between the primary judges in $G_{\text{absent}} - G_{\text{blank}}$ and the encoding estimand the difference between them in $G_{\text{sparse}} - G_{\text{complete}}$, each passing when its two-sided 95\% percentile interval over 10,000 resamples of whole pairs excludes zero and its estimate is at least 5\% in size, with a combined claim requiring both to pass, both to keep passing under the adverse fill of any missing judge verdict, and every comprehension and control criterion of the primary judges to pass. Neither passes (Table~\ref{tab:supp_photo}). The omission-form contrast is $-0.7\%$ $[-2.1\%, +0.7\%]$ and the encoding contrast $-0.2\%$ $[-1.2\%, +0.8\%]$ on 510 pairs, with no judge verdict missing, so the registered bounds equal the estimates; on the 487 screened writer-complete pairs they are $-0.4\%$ $[-1.7\%, +0.9\%]$ and $-0.4\%$ $[-1.3\%, +0.5\%]$, and weighting the 20 categories equally gives $-0.6\%$ $[-2.0\%, +0.8\%]$ and $-0.3\%$ $[-1.3\%, +0.6\%]$. The registered whole-pool diagnostic, which fills the 6 writer-missing pairs at their least and most favourable values, spans $-3.1\%$ $[-5.5\%, -0.9\%]$ to $+1.6\%$ $[-0.5\%, +3.9\%]$ for the omission-form contrast and $-2.4\%$ $[-4.6\%, -0.6\%]$ to $+2.2\%$ $[+0.2\%, +4.5\%]$ for the encoding contrast on 516 pairs, and by registration does not enter the decision. The registered label-error sensitivity adds nothing here, since both estimates already lie below 5\%. The writers state the target object correctly on 94.6\% and 95.4\% of images, and their difference in $Y$ is unresolved under every judge and form, from $-1.2\%$ to $-2.7\%$ under the primary judges; a deterministic reader of the decoded states measures $-2.0\%$ $[-5.1\%, +1.2\%]$, and each primary judge differs from it by at most $0.8\%$ in every form, every such difference unresolved. Writing the omitted objects as not reported lowers $Y$ by $7.5\%$ to $8.7\%$ under both primary judges, the training judge's model by striking denials, 25.2\% and 24.1\% of true denials for the 2 writers, and Gemma by accepting false assertions, 23.3\% and 19.6\%, but for both writers alike. In the checklist style the omission-form effect of every judge lies between $-0.8\%$ and $0.0\%$, none resolved. Phi-4-mini's resolved effects, $+5.8\%$ and $-6.8\%$, come with false acceptance of denials of 46.9\% to 80.2\% on complete and absent-filled records, the failure its comprehension check predicted, and OLMo-2 shows the same failure with unresolved effects. Under the registered result table this outcome leaves the radiology findings as they stand and reports the photograph study as a bounded negative.

\section{Statistics}
\label{sec:supp_stats}

Pairs that share a patient are linked, and the connected groups are the resampling units, 221 on the 238 validation pairs, 220 on the 237 pairs common to the registered replication's endpoint reads, and 245 on the 273 held-out test pairs common to its endpoint reads there. For the radiology analyses every two-sided interval is a 95\% percentile interval from 2,000 bootstrap draws of these groups with a fixed seed (20260925), and paired changes recompute both reads on the same draw. The false-strike condition of the registered criterion uses the one-sided 95\% upper bound of the same bootstrap distribution, which is lower than the upper end of the two-sided interval. The replication averages each pair's change over the 3 seeds before resampling, so its intervals describe patient variation given these 3 trained writers and not variation over further training seeds. The between-reader and between-judge contrasts recompute all quantities on the same draw, so their interval is the interval of the contrast itself. On the development run these contrasts were computed after the corresponding estimates were seen and are post hoc, whereas the endpoint contrasts were registered before the replicated and held-out test records were read, and the between-judge contrast in the full reward-scoring form before the replicated validation records were read. The order-by-fill and target-by-other designs are post hoc for the first replicated run, the not-assessed form is exploratory, and any form analysis of the held-out test records is post hoc. The photograph study resamples whole pairs with 10,000 draws (Sec.~\ref{sec:supp_photo}).

\section{Compute}
\label{sec:supp_compute}

Hours come from the job accounting of the cluster and count every allocated graphics processing unit (GPU). A 40-step registered training run took 4.2 to 5.6 hours on 3 48 GB A6000 GPUs, 12.7 to 16.9 GPU-hours, and an omission-cost run 5.2 to 7.3 hours, since its records are longer. A full read of the 238 validation pairs by the training judge took 0.32 A6000 GPU-hours, and each re-evaluation of saved records by the independent judge or in another form took 3 to 9 minutes on 1 GPU. The 3 replicated runs used 44.8 A6000 GPU-hours, the second writer 14.6, the learning-rate test 23.5, and the pilots before it 48.4, and the direct-supervision runs used 2.6 A100 GPU-hours. The 3 omission-cost runs used 58.2 A6000 GPU-hours, and the re-evaluations by the 3 further judges 1.3 A100 and 2.1 RTX 4090 GPU-hours. Including pilots, check runs, and every read on the main cluster up to 29 September, the study used 279.6 A6000, 12.3 A100, and 10.2 RTX 4090 GPU-hours, plus 10.5 RTX PRO 6000 Blackwell and 4.1 H100 GPU-hours on the Empire AI cluster for development pilots, check runs, and data staging. The registered photograph study's writer and judge jobs used 4.4 A6000 GPU-hours, its 30B judge running on 2 GPUs, excluding its screen, calibration, and check runs. The rereads by judge size and instruction, with their check runs, used 16.3 A6000 GPU-hours.

\section{Data}
\label{sec:supp_data}

Images, reports, and labels come from CheXpert Plus (dataset DOI \url{https://doi.org/10.71718/6nvz-pm34}) under its research use agreement. Dataset files will not be redistributed, the pair definitions will be released as study identifiers, and the figures reproduce de-identified radiographs from the dataset for illustration. The photograph study uses the COCO 2014 images and instance annotations and the VQA v2 questions, answers, and complementary pairs under their Creative Commons Attribution 4.0 annotation licenses and the images' Flickr terms; its pair identifiers, records, verdicts, and frozen protocol will be released.

\section{Figures}
\label{sec:supp_art}

Every figure is drawn by a script from the dataset's radiographs and the study's saved records, verdicts, reasons, and result files.

\end{document}